\documentclass[journal]{IEEEtran}

\usepackage[table,xcdraw]{xcolor}

\usepackage[utf8]{inputenc}

\usepackage{CJK}
\usepackage{float}
\usepackage{amsmath}
\usepackage{amsfonts}
\usepackage{graphicx}
\usepackage{cite}
\usepackage{picinpar}
\usepackage{bigstrut} 
\usepackage{url}
\usepackage{flushend}
\usepackage{array,tabularx}

\newcolumntype{Y}{>{\raggedright\arraybackslash}X}       
\newcolumntype{L}[1]{>{\raggedright\arraybackslash}m{#1}}
\newcolumntype{C}[1]{>{\centering\arraybackslash}m{#1}}  

\usepackage{soul}
\usepackage{pifont}
\usepackage{alltt}
\usepackage{enumerate}
\usepackage{siunitx}
\usepackage{epstopdf}
\usepackage{pbox}
\usepackage{multirow, threeparttable, booktabs, makecell}
\usepackage{diagbox}
\usepackage{bm}
\usepackage{verbatim}
\usepackage[linesnumbered,ruled,vlined]{algorithm2e}
\usepackage{times}
\usepackage{subfigure}
\usepackage{mfirstuc}
\usepackage{textcomp}

\usepackage[hidelinks]{hyperref}

\begin{document}

\title{Distributed Real-Time Vehicle Control for Emergency Vehicle Transit: A Scalable Cooperative Method}

\author{WenXi  Wang,  JunQi Zhang, \IEEEmembership{Senior Member, IEEE}
\thanks{WenXi Wang, JunQi Zhang are with the School of Computer Science and Technology, Key Laboratory of Embedded System and Service Computing, Ministry of Education, Tongji University, Shanghai 200092, China (e-mail: 2410915@tongji.edu.cn; zhangjunqi@tongji.edu.cn)}

\thanks{Corresponding author: JunQi Zhang}
}


\maketitle
\begin{abstract}
Rapid transit of emergency vehicles is critical for saving lives and reducing property loss but often relies on surrounding ordinary vehicles to cooperatively adjust their driving behaviors. It is important to ensure rapid transit of emergency vehicles while minimizing the impact on ordinary vehicles. 
Centralized mathematical solver and reinforcement learning are the state-of-the-art methods. 
The former obtains optimal solutions but is only practical for small-scale scenarios. 
The latter implicitly learns through extensive centralized training but the trained model exhibits limited scalability to different traffic conditions. 
Hence, existing methods suffer from two fundamental limitations: high computational cost and lack of scalability.
To overcome above limitations, this work proposes a scalable distributed vehicle control method, where vehicles adjust their driving behaviors in a distributed manner online using only local instead of global information.
We proved that the proposed distributed method using only local information is approximately equivalent to the one using global information, which enables vehicles to evaluate their candidate states and make approximately optimal decisions in real time without pre-training and with natural adaptability to varying traffic conditions. 
Then, a distributed conflict resolution mechanism is further proposed to 
guarantee vehicles' safety by avoiding their 
decision conflicts, which eliminates the 
single-point-of-failure risk of centralized methods and provides deterministic safety guarantees that learned methods cannot offer.
Compared with existing methods, simulation experiments based on real-world traffic datasets demonstrate that the proposed method achieves faster decision-making, less impact on ordinary vehicles, and maintains much stronger scalability across different traffic densities and road configurations.

\end{abstract}

\begin{IEEEkeywords}
Distributed real-time vehicle control, Rapid transit of emergency vehicles, Traffic impact minimization
\end{IEEEkeywords}

\section{Introduction}
\IEEEPARstart{E}mergency vehicles (EMVs) responding to incidents in transportation networks are crucial for protecting human lives and minimizing property damage\cite{1}. Response time is often used as an important metric to evaluate emergency management capability and system sustainability\cite{3}. For most EMVs, response time refers to the time between receiving an alert and arriving at the incident location. Existing studies have demonstrated that reducing EMV response times enhances disaster response capabilities and minimizes disaster impacts. In medical emergencies, a one-minute reduction in response time can increase survival probability for out-of-hospital cardiac arrest patients by up to 24\% \cite{5}.
In public safety emergencies, a 10\% reduction in response time can increase the likelihood of resolving a crime by 4.7\%\cite{7}. Laws also provide EMVs with special privileges when responding to emergency calls. They have priority on roads and can ignore normal route, direction, speed, and traffic signal restrictions. These privileges speed up EMVs but often disrupt nearby traffic and lower overall transportation system efficiency. Therefore, ensuring efficient and smooth transit of EMVs while minimizing their traffic impact on ordinary vehicles (OVs) has much practical significance. Next, we briefly review traffic control strategies for rapid transit of EMVs and present our contributions.

Traffic control strategies for rapid transit of EMVs can be categorized into the following three types:
1) Route optimization is a strategy to select the optimal routes for EMVs across road networks. The optimization is performed before vehicle dispatch, providing a reference for centralized control systems. This strategy employs various methods, including deterministic algorithms like Dijkstra \cite{9,10,41,42} and heuristic algorithms like particle swarm optimization \cite{11}\cite{12}. Overall, route optimization focuses on EMV routing without considering OV routing. Consequently, the execution result largely depends on the autonomous reactions of the OV drivers, which often leads to uncertainty;
\begin{table*}[htbp]
\centering
\small
\renewcommand{\arraystretch}{1.25}
\setlength{\tabcolsep}{3pt}
\caption{Comparison of Methods Based on Road Segment Vehicle Control Strategy}
\label{tab:comparison}
\begin{tabularx}{\textwidth}{
    >{\centering\arraybackslash}p{1.7cm}
    >{\raggedright\arraybackslash}X
    >{\raggedright\arraybackslash}X
    >{\centering\arraybackslash}X
    >{\centering\arraybackslash}p{2.2cm}
    >{\centering\arraybackslash}p{1.8cm}
    >{\centering\arraybackslash}p{1.4cm}
    >{\centering\arraybackslash}p{1.3cm}
}
\toprule
\textbf{Paper} 
& \textbf{Optimization Objective} 
& \textbf{Solution Method} 
& \textbf{Case Size} 
& \textbf{Algorithm Usage Condition} 
& \textbf{Control Paradigm} 
& \textbf{Training Time} 
& \textbf{Max Lane Count} \\
\midrule
Murray-Tuite \newline et al. 
& EMV speed 
& Genetic algorithm 
& A segment with 20 OVs 
& None 
& Centralized 
& 0 
& 3 \\
\addlinespace[3pt]
Hannoun \newline et al. 
& EMV speed 
& CPLEX 
& A segment with 15 OVs 
& None 
& Centralized 
& 0 
& 3 \\
\addlinespace[3pt]
Wu \newline et al. 
& EMV speed, lane clearance cost 
& A* algorithm \& commercial solvers 
& A segment with 54 OVs 
& None 
& Centralized 
& 0 
& 3 \\
\addlinespace[3pt]
Lin \newline et al. 
& EMV efficiency; state-change count
& Customized algorithm \& Gurobi 
& 3 segments with 144 OVs 
& None 
& Centralized 
& 0 
& 3 \\
\addlinespace[3pt]
Yang \newline et al. 
& EMV efficiency; state-change count
& Graph conv. soft actor-critic (GSAC) 
& A long segment with 80 OVs 
& Within 100\,m ahead of EMV 
& Distributed 
& 10--15 h 
& 3 \\
\addlinespace[3pt]
\rowcolor[HTML]{F5F5F5} 
\textbf{This} \newline \textbf{paper} 
& \textbf{EMV efficiency; state-change count}
& \textbf{Distributed real-time vehicle control}
& \textbf{3 long segments with 436 OVs} 
& \textbf{None} 
& \textbf{Distributed} 
& \textbf{0} 
& \textbf{5} \\
\bottomrule
\end{tabularx}
\end{table*}
2) Intersection signal control is a strategy that adjusts traffic signals to give priority to EMVs. Such control ranges from simple adjustments at individual intersections based on EMV position detection \cite{15}\cite{16}, to coordination of signals between adjacent intersections along rescue routes \cite{17}\cite{18}, and further to network-wide coordination across multiple intersections to maintain overall traffic efficiency \cite{tc1,21,tc2,22}. Since this strategy controls only traffic signals, it faces uncertainty issues in the execution result;
and 3) Road segment vehicle control is a strategy that optimizes objectives by planning trajectories for both EMVs and OVs. Existing methods in this category can be further divided into two types: centralized mathematical solvers and reinforcement learning methods.
For the first type, in early research, Murray-Tuite et al. \cite{23} proposed a mixed-integer nonlinear programming model. They maximized EMV speed as the optimization objective by using a genetic algorithm to determine vehicle states to improve transit efficiency in road segments with 20 OVs. Hannoun et al. \cite{24} improved the model by reformulating it as an integer linear programming model. They also maximized the EMV speed, but used a commercially available solver called CPLEX. However, these methods required OVs to pull over completely. They ensured rapid transit of EMVs but caused significant impact on OVs. Addressing these issues, Wu et al. \cite{25} proposed a flexible lane pre-clearing method that allowed OVs to clear lanes by simply driving forward without pulling over.
However, their method assumes constant speed for all vehicles, which does not reflect the heterogeneous speed distribution in actual traffic conditions.
Moreover, the above studies optimized only for single EMV scenarios. In severe disaster situations, simultaneous dispatching of multiple EMVs is often necessary \cite{26,27,28}. To fill this gap, Lin et al. \cite{30} proposed a cooperative multiple EMV priority model based on actual vehicle environments. Inspired by existing research in vehicle trajectory optimization \cite{31,32,33,34}, they used the speed, acceleration and lane-changing behaviors of both EMVs and OVs as decision variables to achieve cooperative driving among vehicles, while optimizing EMV transit efficiency and minimizing the impact on vehicles. They developed two customized algorithms that reduced the number of decision variables and constraints, enabling them to obtain feasible solutions within acceptable computation time, but remains practical only for small-scale scenarios.
For the second type, Yang et al. \cite{Yang} reformulated above problem as a multi-agent reinforcement learning
task and proposed a Graph Convolutional Soft Actor-Critic (GSAC) 
algorithm. Their method enables each vehicle to independently 
make real-time decisions based on trained models, achieving a 
distributed control paradigm. 
However, this approach has three limitations. First, it has 
restrictive usage conditions: the algorithm is only effective 
for vehicles within 100 $m$ ahead of the EMV, leaving vehicles 
beyond this range unable to receive decision support. Second, 
it requires extensive training time of 10--15 hours before 
deployment. Third, they validated their GSAC only under idealized 
conditions with uniform traffic density and homogeneous OV speeds 
throughout the whole road segment, which differs from real-world 
traffic environments and makes it difficult to demonstrate that 
their method can effectively scale to realistic scenarios. 
Such challenges of high training cost and poor generalization are common for deep reinforcement learning methods in multi-agent systems~\cite{tc3}. 
Table~\ref{tab:comparison} presents a comparison of related studies.

Despite significant advances in road segment vehicle control 
strategies, existing methods suffer from two fundamental 
limitations: high computational cost and lack of scalability. 
Centralized mathematical solvers are only practical for 
small-scale scenarios, while reinforcement learning methods 
require extensive training and exhibit limited adaptability 
to varying traffic conditions.
Addressing these limitations, this work aims to make the following novel contributions:
\begin{enumerate}[1)]
\item It proposes a scalable distributed vehicle control method, where vehicles adjust their driving behaviors in a distributed manner online using only local instead of global information.
We proved that the proposed distributed method using only local information is approximately equivalent to the one using global information, which enables vehicles to evaluate their candidate states and make approximately optimal decisions in real time without pre-training and with natural adaptability to varying traffic conditions. We also propose a distributed conflict resolution mechanism to guarantee vehicles' safety. And
\item It performs simulation experiments based on real-world 
traffic datasets. The results demonstrate that the proposed 
method achieves faster decision-making, less impact on 
ordinary vehicles, and stronger scalability across different 
traffic densities and road configurations.
\end{enumerate}

The remainder of the paper is organized as follows. Section~\ref{sec:related} presents the problem, mathematical formulation used in the state-of-the-art research for rapid transit of emergency vehicles. Section~\ref{sec:method} presents our proposed scalable distributed vehicle control method. Section~\ref{sec:Theoretical} conducts theoretical analysis. Section~\ref{sec:experiments} conducts numerical experiments. Finally, Section~\ref{sec:conclusion} concludes this paper.

\section{RELATED WORK}\label{sec:related}

This section presents the problem, mathematical formulation used in the state-of-the-art research\cite{30} for rapid transit of emergency vehicles, thus laying the foundation for our proposed method.

\begin{figure}
    \includegraphics[width=\columnwidth]{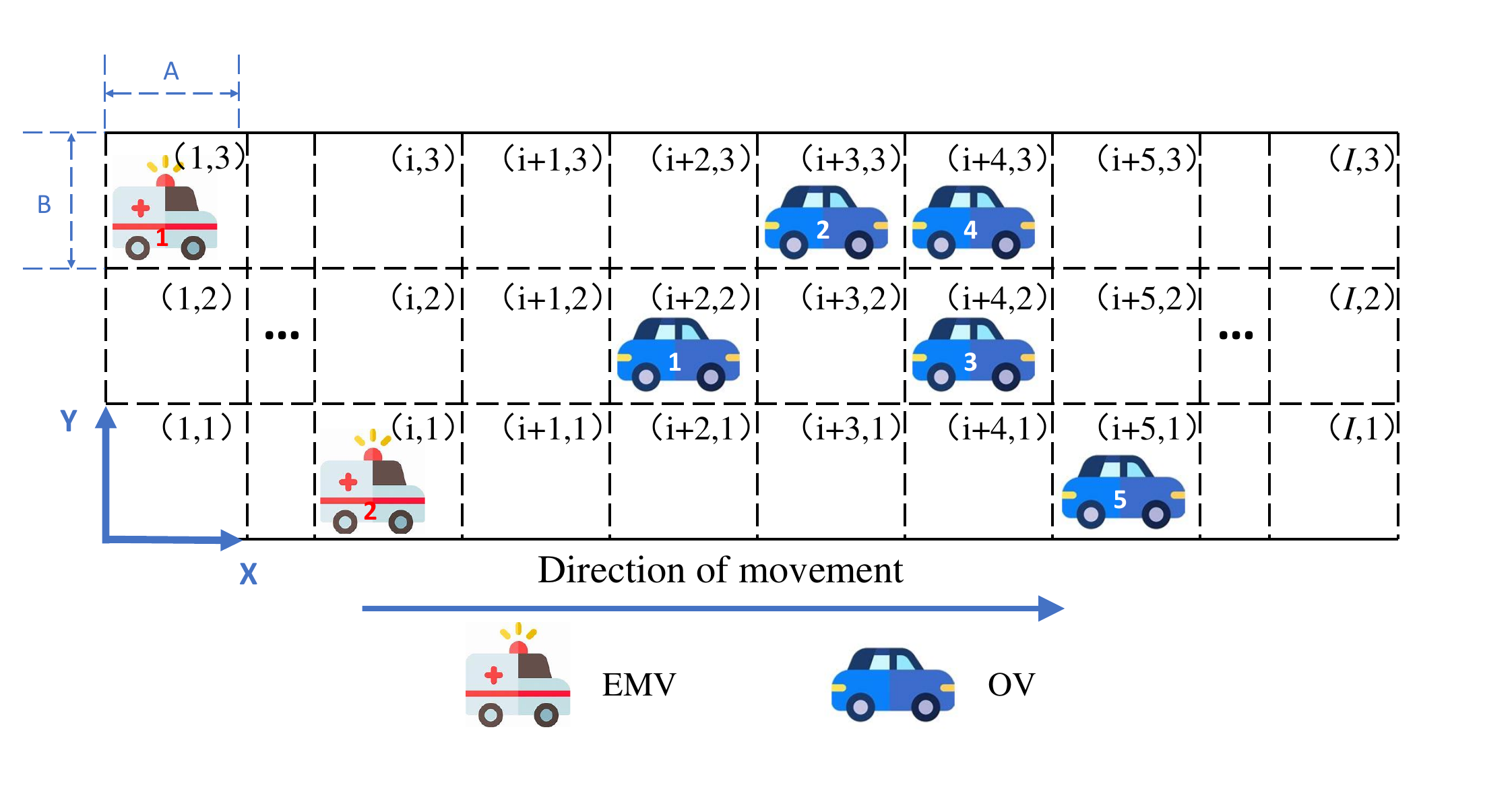}
    \caption{Discretization of the considered roadway network.}
    \label{1}
\end{figure}
\subsection{Problem Statement}

The problem is stated in discretized space. The road segment is discretized into a two-dimensional coordinate system composed of identical grid cells (Fig.~\ref{1}). Each grid cell has a length of $A$ and width of $B$. The $x$-axis of this system represents longitudinal movement, with grid cells labeled from $1$ (where EMVs first appear) and increasing with distance up to $I$, where $I$ represents the considered number of grid cells in a lane. The $y$-axis represents lateral movement, with grid cells labeled from $1$ (bottom lane) to $L$ (top lane), where $L$ exactly matches the actual lane count ($L=3$ in Fig.~\ref{1}). Within this discretized space, $| \mathcal{M}|$ EMVs and $| \mathcal{N}|$ OVs move cooperatively to maximize EMV transit efficiency while minimizing the impact on OVs and EMVs, where the driving strategy of EMVs is to move toward the lane with the lowest traffic density, accelerate to the maximum speed $V_{\max}$ and maintain it. It should be noted that this problem focuses on scenarios where road segments have sufficient space to ensure feasible driving routes for EMVs, while extreme conditions such as traffic jams are beyond the scope of this work\cite{30}. $ \mathcal{M}$ represents the set of EMVs and $ \mathcal{N}$ the set of OVs. $\mathcal{X}$ represents the set of all vehicles, with $\mathcal{M} \subseteq \mathcal{X}$ and $\mathcal{N} \subseteq \mathcal{X}$. EMVs are labeled with $m$ or $m'$ and OVs with $n$ or $n'$. Both EMVs and OVs are indexed separately from left to right, e.g., EMV at (1, 3) is indexed as 1 and EMV at ($i$, 1) as 2 (assuming that $i>1$) in Fig.~\ref{1}. When multiple vehicles are located at the same $x$-coordinate, they are indexed in ascending order from bottom to top, e.g., OV at ($i+4$, 2) is indexed as 3 while OV at ($i+4$, 3) as 4. The vehicle state variables including index, time, position, and speed are all discretized. The speed discretization method is given in Table~\ref{tab:speed_lookup}.

\begin{table}[htbp]
\centering
\caption{SPEED LOOKUP TABLE FOR EMVs AND OVs}
\label{tab:speed_lookup}
\setlength{\tabcolsep}{3pt}  
\begin{tabular*}{\columnwidth}{@{\extracolsep{\fill}}ccc@{}}
\hline
Level & Speed (cells/s) & Speed (m/s) \\
\hline
0 & 0 & 0 \\
1 & 1 & 6 \\
2 & 2 & 12 \\
3 & 3 & 18 \\
4 & 4 & 24 \\
5 & 5 & 30 \\
\hline
\end{tabular*}
\end{table}

\subsection{Problem Formulation}

The dual objectives of this problem are to maximize the transit efficiency for EMVs and minimize the impact on OVs and EMVs. A hierarchical optimization model\cite{35} \cite{36} is constructed to define their priority order. The former enjoys the higher priority than the latter. The transit efficiency is evaluated by calculating the total distance that all EMVs can travel in the considered period. The longer one indicates higher efficiency. The impact on OVs and EMVs is evaluated by calculating the sum of all lane changes of all vehicles and acceleration/deceleration behaviors of OVs in the considered period. Fewer changes indicate lower impact on OVs and EMVs. We use $f$ and $f'$ to represent the maximization of EMVs' transit efficiency and minimization of the impact on OVs and EMVs, respectively:

\begin{gather}
\max_{i^T_{m}} f = \sum_{m \in \mathcal{M}} i^T_{m}, \tag{1} \label{eq:1}\\
\min_{a^{t}_{n}, b^{t}_{n}, \alpha^{t}_{m}, \beta^{t}_{m}, \alpha^{t}_{n}, \beta^{t}_{n}} f^{\prime} = \sum_{t\in \mathcal{T}\backslash\{T\},n} c_1(a^{t}_{n} + b^{t}_{n}) + \nonumber\\
\sum_{t\in \mathcal{T}\backslash\{T\},m} c_2(\alpha^{t}_{m} + \beta^{t}_{m}) + \sum_{t\in \mathcal{T}\backslash\{T\},n} c_3(\alpha^{t}_{n} + \beta^{t}_{n}) \tag{2} \label{eq:2}
\end{gather}
where $\mathcal{T} = \{0, 1, ..., T\}$ represents the set of all time steps; $i^T_{m}$ represents the longitudinal position of EMV $m$ at the final time $T$; $a_{n}^{t}$ and $b_{n}^{t}$ represent the acceleration and deceleration of OV $n$ at time $t$ in $\mathcal{T}$; $\alpha^{t}_{m}$ and $\beta^{t}_{m}$ are binary variables that represent the leftward and rightward lane-changing of EMV $m$ at time $t$; $\alpha^{t}_{n}$ and $\beta^{t}_{n}$ are binary variables that represent the leftward and rightward lane-changing of OV $n$ at time $t$; and $c_1$, $c_2$, and $c_3$ represent weights to balance the importance of three types of behavior changes. Since EMVs follow the predetermined driving strategy, $f$ in (\ref{eq:1}) is implicitly maximized. Thus, the optimization focuses on minimizing $f'$ in (\ref{eq:2}).

\begin{gather}
    v^t_{n} = v^0_{n} + \sum_{0\leq\tau\leq t-1} (a^{\tau}_{n} - b^{\tau}_{n}), \forall t \in \mathcal{T}\backslash\{0\}, n \in \mathcal{N}, \tag{3} \label{eq:3} \\
 i^t_{n} = i^0_{n} + \sum_{0\leq\tau\leq t-1} v^\tau_{n}, \forall t\in\mathcal{T}\backslash\{0\}, n \in \mathcal{N}. \tag{4} \label{eq:4}
\end{gather}

The constraints ensure the continuity and safety of vehicle movement, as well as the traffic efficiency of OVs. (\ref{eq:3}) and (\ref{eq:4}) formulate changes in speed and longitudinal position of OVs. (\ref{eq:3}) describes that the speed of OV $n$ at time $t$ is determined by its initial speed added to the sum of all accelerations and decelerations during that period, where the speed does not exceed the maximum speed $V_{\max}$. (\ref{eq:4}) describes that the longitudinal position of OV $n$ is determined by its initial position added to the sum of the accumulated speed during that period. $v_{n}^t$ represents the speed of OV $n$ at time $t$; $i_{n}^t$ the longitudinal position of OV $n$ at time $t$. 

\begin{gather}
l^t_{m} = l^0_{m} + \sum_{0\leq\tau\leq t-1} (\alpha^{\tau}_{m} - \beta^{\tau}_{m}), \forall t \in \mathcal{T}\backslash\{0\}, m \in \mathcal{M}, \tag{5} \label{eq:5} \\
l^t_{n} = l^0_{n} + \sum_{0\leq\tau\leq t-1} (\alpha^{\tau}_{n} - \beta^{\tau}_{n}), \forall t \in \mathcal{T}\backslash\{0\}, n \in \mathcal{N}. \tag{6} \label{eq:6}
\end{gather}

\eqref{eq:5} and \eqref{eq:6} formulate the changes in lateral position for vehicles. Specifically, they describe that the lateral positions of EMV $m$ and OV $n$ at time $t$ are determined by their initial lateral positions added to the sum of all lane-changing behaviors during that period. $l_{m}^t$ represents the lateral position of EMV $m$ at time $t$; $l^t_{n}$ the lateral position of OV $n$ at time $t$.

To ensure driving safety and prevent vehicle collisions, safety distance constraints are introduced.

\begin{gather}
    d^t_{n} = i^t_{n} + M_1(l^t_{n} - 1), \forall t \in \mathcal{T}, n \in \mathcal{N}, \tag{7} \label{eq:7} \\
d^t_{n'} - d^t_{n} \geq M_2(w_{t,n',n} - 1) + v^t_{n} - v^t_{n'} + 1, \nonumber \\
\forall t \in \mathcal{T}, n' \neq n \in \mathcal{N}, \tag{8} \label{eq:8} \\
d^t_{n'} - d^t_{n} \leq M_2 w_{t,n',n} - 1, \forall t \in \mathcal{T}, n' \neq n \in \mathcal{N}. \tag{9} \label{eq:9}
\end{gather}

\eqref{eq:7} formulates the one-dimensional coordinate position of OVs, which integrates lateral position and longitudinal one. $M_1$ represents the weight for lateral position in this one-dimensional coordinate. \eqref{eq:8} and \eqref{eq:9} formulate distance constraints among OVs. $M_2$ serves as the weight that controls the influence of $w_{t,n',n}$ on the constraints. \eqref{eq:8} describes the minimum safety distance that must be maintained among OVs. \eqref{eq:9} describes the maximum distance constraint among OVs, ensuring that the distance among OVs is limited within the considered grid cells.
$d_{n}^t$ represents the one-dimensional coordinate of OV $n$ at time $t$. $w_{t,n',n}$ is a binary variable that is 1 if $d_{n'}^t > d_{n}^t$ at time $t$ and 0 otherwise.  $M_1$ and $M_2$ are both large constants. It is recommended that their values be chosen as $I$ and $3I$, respectively, as in \cite{30}.

Similarly, the safety distance constraints that must be maintained among EMVs and OVs are as follows:

\begin{gather}
d^t_{m} = (i^0_{m} + \sum_{0\leq\tau\leq t-1} v^{\tau}_{m}) + M_1(l^t_{m} - 1), \nonumber \\
\forall t \in \mathcal{T}\backslash\{0, T\}, m \in \mathcal{M}, \tag{10} \label{eq:11} \\[4pt]
d^t_{n} - d^t_{m} \geq M_2(\phi_{t,m,n} - 1) + v^t_{m} - v^t_{n} + 1, \nonumber \\
\forall t \in \mathcal{T}, m \in \mathcal{M}, n \in \mathcal{N}, \tag{11} \label{eq:12} \\[4pt]
d^t_{n} - d^t_{m} \leq M_2\phi_{t,m,n} - 1, \forall t \in \mathcal{T}, m \in \mathcal{M}, n \in \mathcal{N} \tag{12} \label{eq:13} 
\end{gather}
where $d^t_{m}$ represents the one-dimensional coordinate of EMV $m$ at time $t$. $\phi_{t,m,n}$ is a binary variable that is 1 if $d_{n}^t > d^t_{m}$ at time $t$ and 0 otherwise.

\begin{gather}
v^T_{n} \geq \min\{v^0_{n}, \bar{V}_{\text{OV}}\}. \tag{13} \label{eq:14}
\end{gather}

To ensure the traffic efficiency of OVs, \eqref{eq:14} formulates a constraint on the speed of OVs at the final time, ensuring that their speed remains above a specific threshold (the minimum of the initial speed and the mean speed $\bar{V}_{\text{OV}}$ of the traffic flow at the initial time).

\section{SCALABLE DISTRIBUTED VEHICLE CONTROL}\label{sec:method}

This section presents a Scalable Distributed Vehicle Control method for rapid transit of emergency vehicles (SDVC). Specifically, each OV first judges whether it is influenced by surrounding vehicles. When not influenced, the OV maintains its current driving state. Otherwise, it evaluates all feasible next states using the strategy function and selects the optimal one as its candidate next state. Finally, SDVC resolves conflicts through coalition formation.

\subsection{Influence Judgment}

\begin{figure}[t]
\centering
\subfigure[When vehicle $j$ is an EMV or cooperatively driving with the EMV, it cannot decelerate, so OV $n$ requires a longer prediction horizon.]{
    \includegraphics[width=\columnwidth,trim=6 35 6 25,clip]{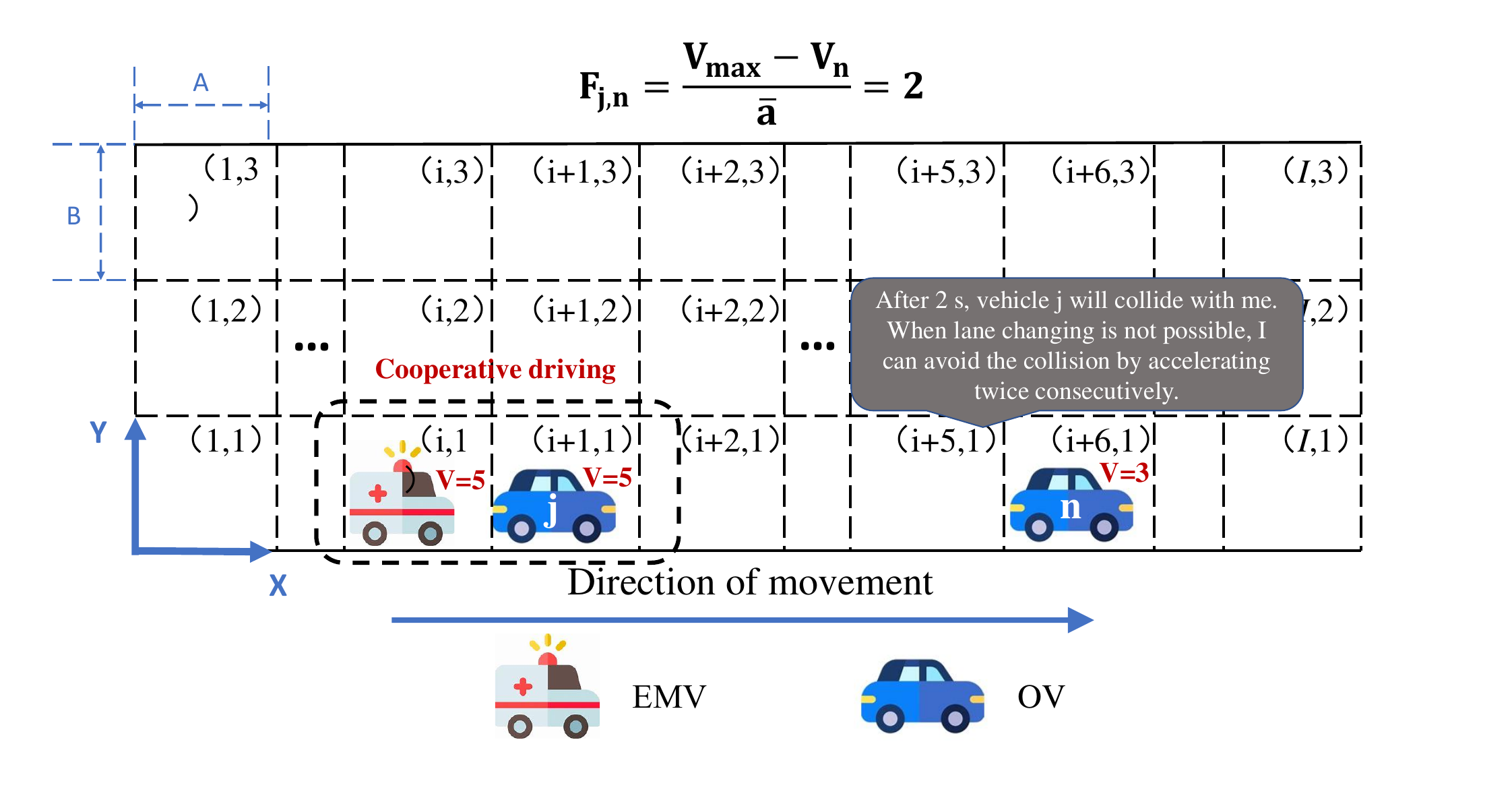}
    \label{fig:F1}
}
\subfigure[When both vehicles are OVs, both can adjust their speeds, allowing a shorter prediction horizon.]{
    \includegraphics[width=\columnwidth,trim=6 40 4 25,clip]{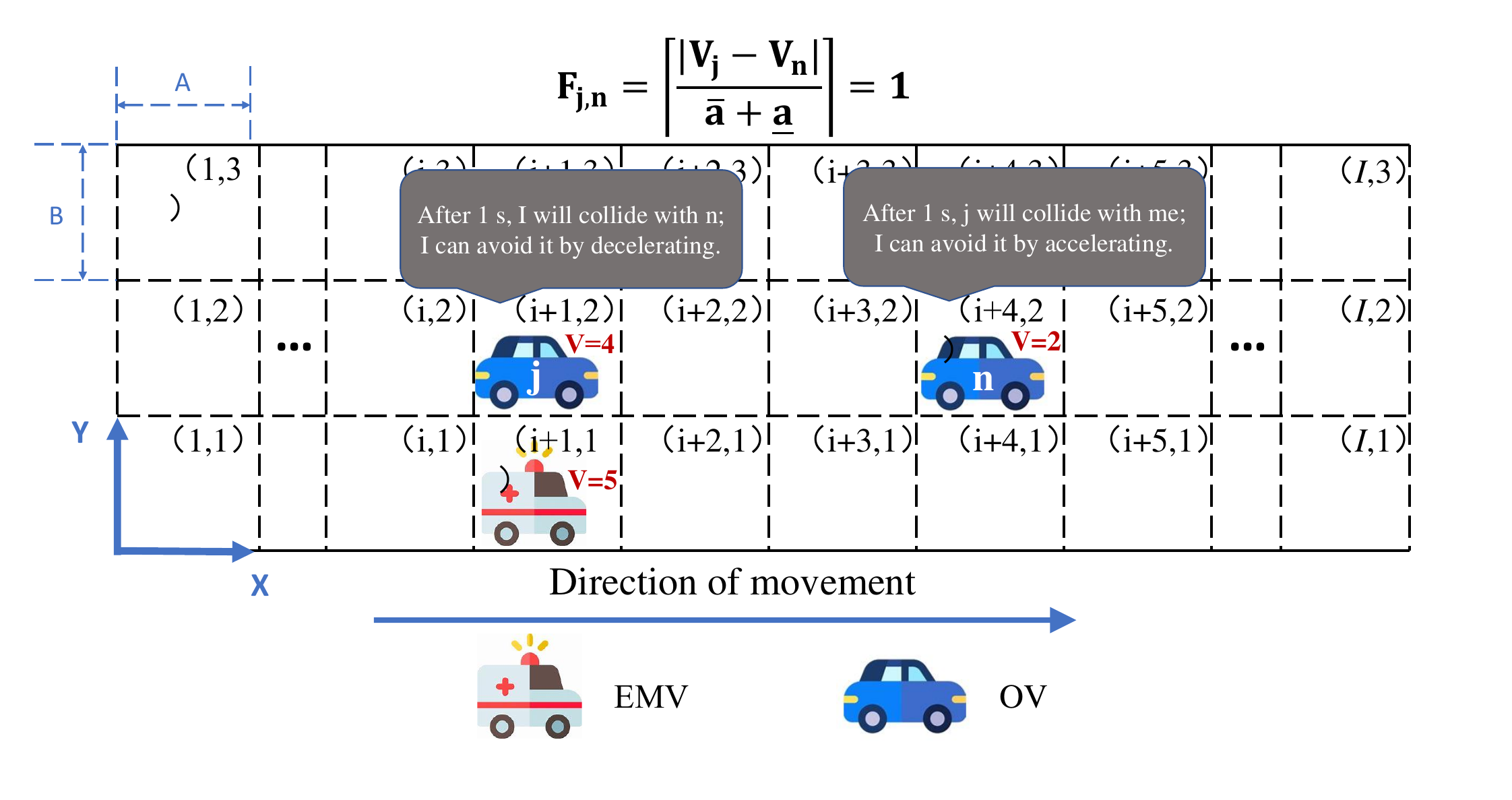}
    \label{fig:F2}
}
\caption{Illustration of prediction horizon $F_{j,n}$.}
\label{fig:prediction_horizon}
\end{figure}

Let $\mathcal{C}_n$ denote the set of vehicles within the communication range of OV $n$.
For each neighbor $j\in\mathcal{C}_n$, OV $n$ assigns a prediction horizon $F_{j,n}$ based on the speed difference between them.
If $j$ is an EMV or an OV cooperatively driving with the EMV, since they cannot decelerate, $F_{j,n} \geq (V_{\max}-v_n^t)/\bar{a}$ to ensure collision avoidance (see Fig.~\ref{fig:F1}).
For other OVs, $F_{j,n} \geq \left\lceil |v_j^t-v_n^t|/(\bar{a}+\underline{a}) \right\rceil$ since both vehicles can adjust their speeds (see Fig.~\ref{fig:F2}).
$\bar{a}$ and $\underline{a}$ denote the maximum acceleration and deceleration, respectively.

Based on the prediction horizon $F_{j,n}$, OV $n$ predicts the future states of surrounding vehicles. For each $j\in\mathcal{C}_n$, the predicted next-state sequence is denoted as $\{\check{s}^{t+\tau}_{j,n}\}_{\tau=1}^{F_{j,n}}$, where $\check{s}^{t+\tau}_{j,n} = (\check{i}^{t+\tau}_{j,n}, \check{v}^{t+\tau}_{j,n}, \check{l}^{t+\tau}_{j,n})$ represents the predicted longitudinal position, speed, and lane of vehicle $j$ at time $t+\tau$ from OV $n$'s perspective.

For an EMV $m\in\mathcal{C}_n$, OV $n$ predicts that the EMV accelerates toward $V_{\max}$ and moves to the lane with the lowest traffic density. The predicted states satisfy:
\begin{gather}
\check{i}^{t+\tau+1}_{m,n}
=
\check{i}^{t+\tau}_{m,n}
+
\check{v}^{t+\tau}_{m,n},
\tag{14}\label{eq:15}\\[2pt]
\check{v}^{t+\tau+1}_{m,n}
=
\min\!\left(
\check{v}^{t+\tau}_{m,n}
+\bar{a},\,
V_{\max}
\right),
\tag{15}\label{eq:16}\\[2pt]
\check{l}^{t+\tau+1}_{m,n}
=
\check{l}^{t+\tau}_{m,n}
+
\max\!\Big(
-1,
\min\big(
1,
\check{\lambda}_{m,n}
-
\check{l}^{t+\tau}_{m,n}
\big)
\Big)
\tag{16}\label{eq:17}
\end{gather}
where the initial conditions are $\check{i}^{t}_{m,n}=i_m^t$, $\check{v}^{t}_{m,n}=v_m^t$, $\check{l}^{t}_{m,n}=l_m^t$, and $\tau=0,\dots,F_{m,n}-1$. $K_{l,n}$ denotes the traffic density of lane $l$ within the communication range of OV $n$, and $\check{\lambda}_{m,n} = \arg\min_{l\in\mathcal{L}} K_{l,n}$ is the EMV's target lane with minimum traffic density.

For an OV $j\in\mathcal{C}_n$, OV $n$ predicts that it maintains its current lane and moves according to its current motion state:
\begin{gather}
\check{i}^{t+\tau}_{j,n}
=
i_{j}^t
+
\tau\,v_{j}^t,
\quad
\check{v}^{t+\tau}_{j,n}
=
v_{j}^t,
\quad
\check{l}^{t+\tau}_{j,n}
=
l_{j}^t
\tag{17}\label{eq:18}
\end{gather}
where $\tau=1,\dots,F_{j,n}$.

Similarly, OV $n$ moves according to its current motion state, maintaining constant speed and lane:
\begin{gather}
\check{i}^{t+\tau}_{n}
=
i_{n}^t
+
\tau\,v_{n}^t,
\quad
\check{v}^{t+\tau}_{n}
=
v_{n}^t,
\quad
\check{l}^{t+\tau}_{n}
=
l_{n}^t
\tag{18}\label{eq:19}
\end{gather}
where $\tau=1,\dots,F_{j,n}$.

\begin{figure}[t]
\centering
\includegraphics[width=\columnwidth,trim=6 30 6 35,clip]{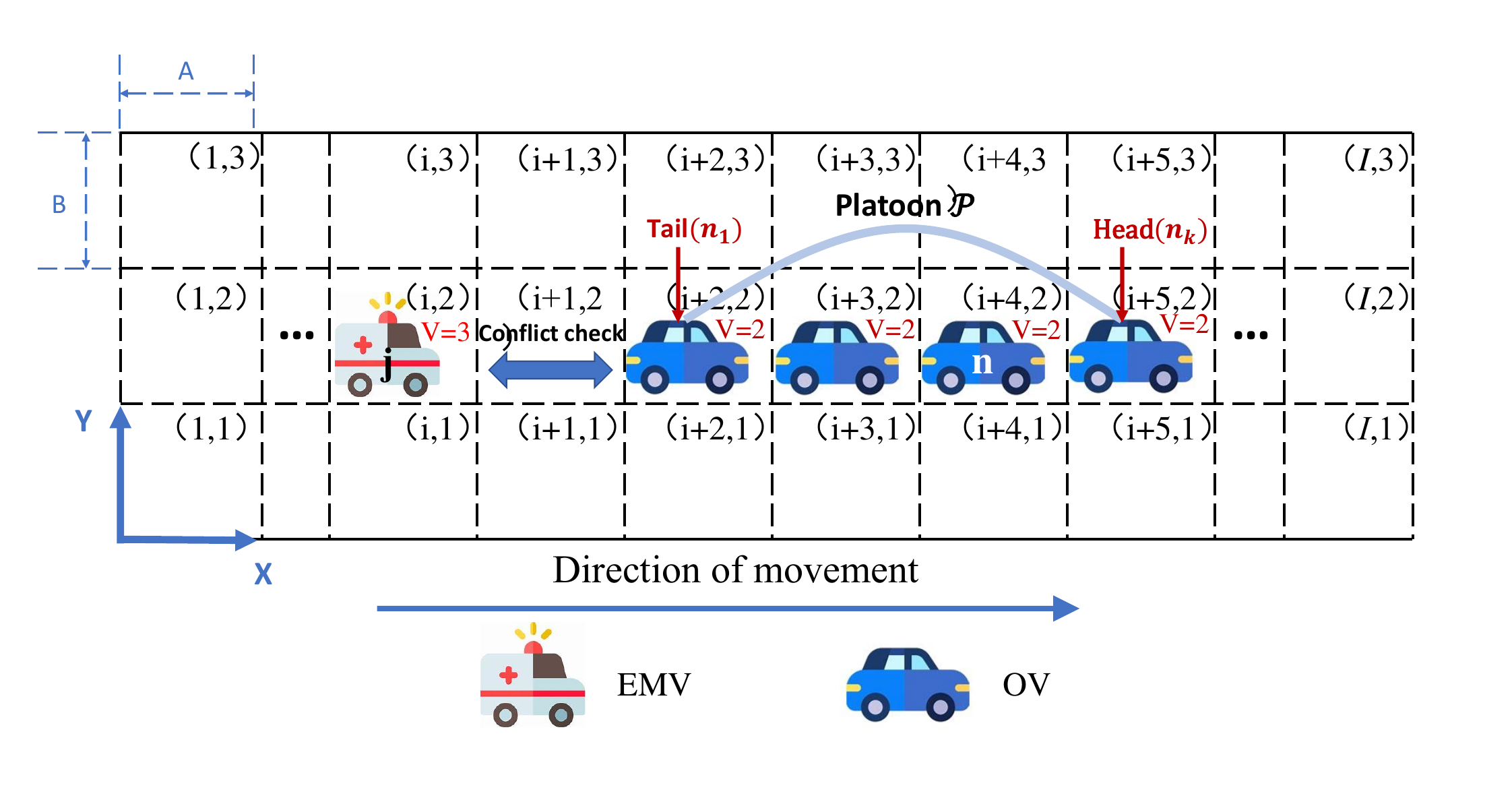}
\caption{Illustration of a platoon. Vehicles in the platoon share the same lane and speed with consecutive positions. OV $n$ needs to check whether the platoon's head ($n_k$) or tail ($n_1$) conflicts with other vehicles.}
\label{fig:platoon}
\end{figure}

Based on the predicted states, OV $n$ is judged to be influenced when both of the following conditions are satisfied:

\textbf{Condition 1 (Safety Violation):}
The predicted states of OV $n$ and some vehicles violate the safety constraints (\ref{eq:7})--(\ref{eq:9}) or (\ref{eq:11})--(\ref{eq:13}). Alternatively, when OV $n$ is part of a platoon formed by multiple vehicles (see Fig.~\ref{fig:platoon}), OV $n$ checks whether its platoon's head or tail violates the safety constraints with other vehicles. This design ensures that all vehicles in the platoon perceive the influence simultaneously and respond cooperatively.

\textbf{Condition 2 (Speed Deviation):}
OV $n$'s speed deviates more from the lane mean speed than vehicle $j$'s speed:
\begin{gather}
\left|
v^{t}_{n}
-
\bar{v}_{{l}^{t}_{n}}
\right|
>
\left|
v^{t}_{j}
-
\bar{v}_{{l}^{t}_{n}}
\right|,
\quad
j \in \mathcal{C}_n
\tag{19}\label{eq:20}
\end{gather}
where $\bar{v}_{{l}^{t}_{n}}$ is the mean speed of lane ${l}^{t}_{n}$. This design ensures normal driving for most vehicles whose speed is close to $\bar{v}_{l^t_n}$, and requires only a few vehicles to change their behavior, effectively minimizing overall behavior changes for OVs. Algorithm \ref{alg:influence_judgment} shows the pseudocode for the influence judgment.

The lane mean speed $\bar{v}_{{l}^{t}_{n}}$ is calculated as:
\begin{gather}
\bar{v}_{{l}^{t}_{n}}
=
\begin{cases}
V_{\max},
& \text{if } \exists\, m \in \mathcal{M}:\,
{l}^{t}_{n}=\check{\lambda}_{m,n}
\wedge
{i}^{t}_{m}<{i}^{t}_{n}
\\[6pt]
\dfrac{1}{|\mathcal{X}_{{l}^{t}_n}|}
\displaystyle\sum_{j \in \mathcal{X}_{{l}^{t}_n}}
v^{t}_{j},
& \text{otherwise}
\end{cases}
\tag{20}\label{eq:21}
\end{gather}
where $\mathcal{M}$ is the set of EMVs, and $\mathcal{X}_{{l}^{t}_{n}} \subseteq \mathcal{C}_n$ is the set of vehicles currently in lane ${l}^{t}_{n}$. If an EMV is in the lane ${l}^{t}_{n}$ and behind OV $n$, the lane mean speed is set to $V_{\max}$. This setting effectively prevents other vehicles from changing to this lane and obstructing the EMV's transit. Otherwise, the lane mean speed is calculated as the average speed of all vehicles in that lane.

\begin{algorithm} 
    \SetAlgoLined
    \caption{Influence Judgement for OV $n$}
    \label{alg:influence_judgment}
\KwIn{$s^{t}_{n}$, $\{s^{t}_{j} | j \in \mathcal{C}_n\}$}
\KwOut{Boolean}
    Identify platoon $\mathcal{P}_n$ containing OV $n$\;
    $(n_{\text{head}}, n_{\text{tail}}) \leftarrow$ Head and tail of $\mathcal{P}_n$ (or $(n, n)$ if no platoon)\;
    \For{\textnormal{each} $j \in \mathcal{C}_n$}
    {     
        Compute prediction horizon $F_{j,n}$\;
        \eIf{$j \in \mathcal{M}$}
        {
            $\{\check{s}^{t+\tau}_{j,n}\}_{\tau=1}^{F_{j,n}} \leftarrow$ Predict using (\ref{eq:15})--(\ref{eq:17})\;
            safeConstraints $\leftarrow$ (\ref{eq:11})--(\ref{eq:13})\;
        }
        {
            $\{\check{s}^{t+\tau}_{j,n}\}_{\tau=1}^{F_{j,n}} \leftarrow$ Predict using (\ref{eq:18})\;
            safeConstraints $\leftarrow$ (\ref{eq:7})--(\ref{eq:9})\;
        }
        $n_{\text{check}} \leftarrow n_{\text{tail}}$ \textbf{if} $i^t_j < i^t_{n_{\text{tail}}}$ \textbf{else} $n_{\text{head}}$\;
        $\{\check{s}^{t+\tau}_{n_{\text{check}}}\}_{\tau=1}^{F_{j,n}} \leftarrow$ Predict using (\ref{eq:18}) or (\ref{eq:19})\;
        \For{$\tau = 1$ \KwTo $F_{j,n}$}
        {
            \If{$(\check{s}^{t+\tau}_{n_{\text{check}}}, \check{s}^{t+\tau}_{j,n})$ \textnormal{violates safeConstraints}}
            {
                \If{\textnormal{satisfies condition (\ref{eq:20})}}
                {
                    \Return TRUE\;
                }
            }
        }
    } 
    \Return FALSE\;
\end{algorithm}

\subsection{Strategy Function}
When an OV is influenced, it utilizes the strategy function to evaluate all feasible next states and selects the one with the best value as its candidate next state. The feasible next states are determined by the following constraints: the speed must satisfy $\max(v^t_{n}-\underline{a},0) \leq v^{t+1}_{n} \leq \min(v^t_{n}+\bar{a},V_{max})$, the longitudinal position must satisfy $i^{t+1}_{n} = i^t_{n}+v^t_{n}$; the lateral position must satisfy $\max(l^t_{n}-1,1) \leq l^{t+1}_{n} \leq \min(l^t_{n}+1,L)$, which is consistent with \cite{30}. 

\begin{gather}
    F_n = w_1 \cdot f_1 + w_2 \cdot f_2 + w_3 \cdot f_3. \tag{21}\label{eq:25}
\end{gather}

To minimize the optimization objective (\ref{eq:2}), the designed strategy function consists of three parts, as shown in (\ref{eq:25}). 
$F_n$ represents the strategy function of OV $n$; $w_1$, $w_2$, and $w_3$ the weights of three parts.

\begin{gather}
    f_1 = c_1\left|v^{t+1}_{n} - v^t_{n}\right| +c_3 \left|l^{t+1}_{n} - l^t_{n}\right|. \tag{22} \label{eq:26}
\end{gather}

It represents OV $n$'s own behavior changes. It should make minimal behavior changes at each time step. (\ref{eq:26}) describes its own behavior changes by summing the absolute values of lane and speed changes.

\begin{gather}
    f_2 = \left|v^{t+1}_{n} - \bar{v}_{l^{t+1}_{n}}\right|. \tag{23} \label{eq:27}
\end{gather}

It represents the speed deviation between OV $n$ and its target lane. The intuition behind this term is that when a vehicle's speed differs significantly from the lane mean speed, it is more likely to trigger behavior changes in surrounding vehicles during subsequent driving, thereby increasing the impact on other vehicles. Minimizing $f_2$ encourages vehicles to select next states where the vehicle's speed $v^{t+1}_{n}$ is close to $\bar{v}_{l^{t+1}_{n}}$ , reducing future behavior adjustments.
\begin{gather}
f_3 = \begin{cases}
1, & \text{if } (\exists j \in \mathcal{C}_n \setminus \mathcal{P}_n, \ s^{t+1}_{n} \text{ and } \check{s}^{t+1}_{j,n} \text{ violate safety}\\
& \text{distance constraints) or } v^{t+1}_{n} < \min\{v^0_{n}, \bar{V}_{\text{OV}}\}, \\
0, & \text{otherwise.}
\end{cases} \tag{24} \label{eq:28}
\end{gather}

$f_3$ represents the penalty. To ensure safety distance constraints among vehicles, i.e., (\ref{eq:7})--(\ref{eq:9}) or (\ref{eq:11})--(\ref{eq:13}), as well as efficiency constraint (\ref{eq:14}), (\ref{eq:28}) describes a penalty mechanism. If the vehicle's next state violates safety distance constraints with any vehicle's predicted next state, or when the vehicle's next state violates the efficiency constraint, $f_3$ is set to 1, indicating that a penalty is assigned. Otherwise, $f_3 = 0$, indicating that no penalty is given. Note that vehicles within the same platoon $\mathcal{P}_n$ are excluded from the safety check, since they are influenced simultaneously with OV $n$ and select their next states using the strategy function, making their next states unpredictable by (\ref{eq:18}); instead, safety among platoon members is guaranteed through the conflict resolution mechanism.

Based on the evaluation for all feasible next states, the vehicle prioritizes selecting the state with the lowest value as its candidate next state. When multiple states have the same lowest value, from the consideration of vehicle safety\cite{37}, the vehicle prioritizes selecting the state that maintains the current lane; if no such state exists, the next state is randomly selected from all the candidate next ones.

\subsection{Conflict Resolution}

After obtaining candidate next states, vehicles synchronize this information through vehicle-to-vehicle (V2V) communication, which has been widely adopted in cooperative driving of connected vehicles~\cite{tc4,tc5}. Since candidate next states may conflict (i.e., violating safety distance constraints), we propose a conflict resolution mechanism based on coalition formation.
This mechanism includes two main steps: conflict coalition initialization and priority-based conflict resolution.
For OV $n$, the process first adds itself to the coalition, then checks non-coalition vehicles whose candidate next states violate safety distance constraints with its own and adds those vehicles to the coalition. For each newly added vehicle, similar checks are conducted. The coalition formation process terminates when no new vehicle is available to join the coalition. The final coalition represents a group of vehicles that require coordinated adjustments to their next states. The conflict coalition is a dynamically expanding set with a maximum capacity of $|\mathcal{C}_n|$ for OV $n$. This expansion method captures all transitively related conflicts, ensuring that each vehicle is assigned to exactly one coalition at each decision step, thereby avoiding duplicate processing and inconsistent decisions.
Algorithm \ref{alg:build_conflict_coalition} shows the pseudocode for conflict coalition initialization.

\begin{algorithm}
    \SetAlgoLined
    \caption{\parbox{10cm}{Conflict Coalition Initialization for OV $n$}}
    \label{alg:build_conflict_coalition}
    \KwIn{$s^{t+1}_{n}$, $\{s^{t+1}_{j} | j \in \mathcal{C}_n\}$}
    \KwOut{Conflict Coalition $G_n$}
    
    Initialize $G_n \leftarrow \{n\}$\;
    new\_vehicle $\leftarrow$ TRUE\;
    
    \While{\textnormal{new\_vehicle and $|G_n| < |C_n|$}}{
        new\_vehicle $\leftarrow$ FALSE\;
        $G_{\text{temp}} \leftarrow \emptyset$\;
        
        \For{\textnormal{each $j \in \mathcal{C}_n \setminus G_n$}}{
            \For{\textnormal{each $i \in G_n$}}{
                \If{\textnormal{$s^{t+1}_{i}$ and $s^{t+1}_{j}$ violate safety distance constraints}}{
                    $G_{\text{temp}}$.append($j$)\;
                    new\_vehicle $\leftarrow$ TRUE\;
                    break\;
                }
            }
        }
        
        $G_n \leftarrow G_n \cup G_{\text{temp}}$\;
    }
    
    \Return{$G_n$}\;
\end{algorithm}

After conflict coalition initialization, conflicts are resolved through a priority-based method. A priority order is established based on the vehicle type and the number of feasible solutions: EMVs have the highest priority, while OVs are prioritized in ascending order of feasible solution counts. When multiple OVs share identical feasible solution counts, a small random perturbation $\epsilon \sim U(-0.5, 0.5)$ is applied to break ties.
The highest-priority OV in the coalition serves as the central vehicle, responsible for computing next states for all coalition members; other OVs wait to receive their assigned next states from the central vehicle. According to their priorities, for each OV, the strategy function evaluates candidate states by considering only vehicles outside the coalition and coalition members with higher priority.
After decisions are made for all vehicles, if conflicts persist, the coalition is expanded by including the nearest vehicle to increase the solution space. 
\begin{gather}
    \overset{\circ}{v} = \min_{i \in \mathcal{C}_n \setminus G_n} \sum_{j \in G_n} d(i, j). \tag{25} \label{eq:29}
\end{gather}

\eqref{eq:29} identifies this vehicle, where $\overset{\circ}{v}$ represents the nearest vehicle and $d(i,j)$ denotes the Manhattan distance between vehicle $i$ and coalition vehicle $j$. The expansion process continues iteratively until conflicts are resolved or the coalition size reaches the maximum communication capacity $|\mathcal{C}_n|$.
If conflicts remain unresolved when the coalition reaches its maximum size, the algorithm terminates and adopts the solution with the fewest conflicts. Such cases only occur under extreme congestion with limited feasible positions, which is beyond the scope as stated in Section \ref{sec:related}.
Algorithm~\ref{alg:priority_resolution} provides the pseudocode for priority-based conflict resolution, and Algorithm~\ref{alg:distributed_control} shows the complete SDVC process.

\begin{algorithm}
    \SetAlgoLined
    \caption{\parbox{10cm}{Priority-based Conflict Resolution for OV $n$}}
    \label{alg:priority_resolution}
    \KwIn{Conflict coalition $G_n$}
    \KwOut{Next state $s^{t+1}_{n}$}
    $numFeasible \leftarrow \{\}$\;
    \For{\textnormal{each vehicle $i \in G_n$}}{
    $numFeasible[i] \leftarrow$ calculate using (\ref{eq:25})\;
    }
    $c_v$ is the highest priority OV in $G_n$\;
    \If{\textnormal{$n$ == $c_v$}}{
    \While{\textnormal{ $|G_n| \leq C_n$}}{
            Sort $G_n$ by vehicle type and $numFeasible$\;
            \For{\textnormal{each vehicle $i \in G_n$ in priority order}}{
                \If{\textnormal{$i$ is EMV}}{
                    $s^{t+1}_{i} \leftarrow$ maintain initial next state\;
                }
                \Else{
                    $s^{t+1}_{i} \leftarrow$ using (\ref{eq:25}) based on priority\;
                }
            }
            
            hasConflicts $\leftarrow$ FALSE\;
            
            \For{\textnormal{$i, j \in G_n, i \neq j$}}{
                \If{\textnormal{$s^{t+1}_{i}$ and $s^{t+1}_{j}$ violate safety distance constraints}}{
                    $\overset{\circ}{v} \leftarrow$ the nearest vehicle using (\ref{eq:29})\;
                    $numFeasible[\overset{\circ}{v}] \leftarrow$ using (\ref{eq:25})\;
                    $G_n$.append($\overset{\circ}{v}$)\;
                    hasConflicts $\leftarrow$ TRUE\;
                    break\;
                }
            }
            
            \If{\textnormal{hasConflicts = FALSE}}{
                Broadcast next states to $G_n$\;
                break;
            }
        }
        
    }
    \Else{
            Wait for next state $s^{t+1}_{n}$ from $c_v$\;
        }
    \Return{$s^{t+1}_{n}$}\;
\end{algorithm}

\begin{algorithm}
    \SetAlgoLined
    \caption{\parbox{10cm}{SDVC for OV $n$}}
    \label{alg:distributed_control}
    \KwIn{$s^{t}_{n}$ and $\{s^{t}_{j} | j \in \mathcal{C}_n\}$}
    \KwOut{$s^{t+1}_{n}$}
    
    $s^{t+1}_{n} \leftarrow (i^t_{n} + v^t_{n}, v^t_{n}, l^t_{n})$\;
    
    influenced $\leftarrow$ Algorithm \ref{alg:influence_judgment} for influence judgment\;
    
    \If{\textnormal{influenced}}{
        $s^{t+1}_{n} \leftarrow$ determine the next state using (\ref{eq:25})\;
    }
    
Exchange information with vehicles in $\mathcal{C}_n$\;
$G_n \leftarrow$ Algorithm \ref{alg:build_conflict_coalition} for conflict coalition formation\;
    
    \If{\textnormal{$|G_n| > 1$}}{
        $s^{t+1}_{n} \leftarrow$ Algorithm \ref{alg:priority_resolution} for conflict resolution\;
    }
    
    \Return{$s^{t+1}_{n}$}\;
\end{algorithm}

\section{THEORETICAL ANALYSIS}\label{sec:Theoretical}

\subsection{Computational Complexity}

The SDVC method consists of three parts: influence judgment, evaluation of strategy function, and conflict resolution.

For the influence judgment part (Algorithm~\ref{alg:influence_judgment}), the loop in Lines 3--21 iterates over each vehicle $j \in \mathcal{C}_n$, where the prediction horizon $F_{j,n} \leq V_{\max}/\underline{a}$ is bounded. Therefore, the time complexity of influence judgment is $O(|\mathcal{C}_n|)$.

For the evaluation of strategy function part, OV $n$ evaluates at most $3 \cdot (\bar{a}+\underline{a}+1)$ feasible next states, where evaluating $f_3$ in~\eqref{eq:28} requires checking safety distance constraints with all vehicles in $\mathcal{C}_n$. Therefore, the time complexity of strategy function evaluation is $O(|\mathcal{C}_n|)$.

For the conflict resolution part (Algorithms~\ref{alg:build_conflict_coalition} and~\ref{alg:priority_resolution}), both phases share the maximum coalition capacity $|\mathcal{C}_n|$. In Algorithm~\ref{alg:build_conflict_coalition}, Lines 6--14 check safety constraints with complexity $O((|\mathcal{C}_n| - |G_n|) \times |G_n|)$. In Algorithm~\ref{alg:priority_resolution}, Line 8 performs sorting in $O(|G_n|\log|G_n|)$, Lines 9--16 adjust states in $O(|G_n| \times |\mathcal{C}_n|)$, and Lines 18--26 check conflicts in $O(|G_n|^2)$. In the worst case where the coalition expands to its maximum capacity $|\mathcal{C}_n|$, the time complexity of conflict resolution simplifies to $O(|\mathcal{C}_n|^3)$.

The worst-case time complexity of SDVC is $O(|\mathcal{C}_n|^3)$, determined by the conflict resolution part. Since $|\mathcal{C}_n|$ represents only the number of vehicles within the communication range rather than the total number of vehicles, the computation time is independent of the problem scale, providing significant scalability advantages.

\subsection{Rationales of the Strategy Function}

We show that minimizing the strategy function of each vehicle is approximately equivalent 
to minimizing its expected contribution to the global optimization objective~\eqref{eq:2}, 
and thus approximately minimizes the expected value of the optimization objective itself.

\subsubsection{Optimization Objective Decomposition}

The optimization objective~\eqref{eq:2} can be decomposed into per-vehicle, per-time-step costs as
\begin{gather}
    f^{\prime} 
    = \sum_{t \in \mathcal{T} \setminus \{T\}} \sum_{n \in \mathcal{N}} g^t_n + C_0
    \tag{26}\label{eq:30}
\end{gather}
where $g^t_n$ denotes the instantaneous cost of OV $n$ at time $t$, and $C_0$ is the total cost associated with EMVs.

According to~\eqref{eq:2}, the instantaneous cost of OV $n$ at time $t$ is
\begin{gather}
    g^t_n 
    = c_1 \left|v^{t+1}_{n} - v^t_{n}\right| +c_3\left|l^{t+1}_{n} - l^t_{n}\right|.
    \tag{27}\label{eq:31}
\end{gather}

\subsubsection{Vehicle Contribution to the Optimization Objective}

From the perspective of OV $n$, when it chooses a next state $s$ at time $t$, its choice affects $f'$ through a local contribution. 
Specifically, its contribution to the optimization objective $f'$ consists of two parts:
\begin{gather}
    \text{Contribution}_n(t \mid s) 
    = g^t_n(s) + \text{FCC}_n(t \mid s)
    \tag{28}\label{eq:32}
\end{gather}
where $g^t_n(s)$ is the instantaneous cost given by~\eqref{eq:31}, and $\text{FCC}_n(t \mid s)$ denotes the future coordination cost due to choosing state $s$ at time $t$.

\subsubsection{Approximation of Future Coordination Cost}

Following the quadratic penalty approach for disagreements in~\cite{^22}, we approximate the future coordination cost that OV $n$ incurs by choosing state $s$ as a weighted sum of squared speed differences with its same-lane neighbors. Suppose OV $n$ coordinates with $K$ vehicles in the same lane, whose
speeds at time $t+1$ are denoted by $v_1^{t+1}, \ldots, v_K^{t+1}$. The future
coordination cost is modeled as:
\begin{gather}
    \sum_{k=1}^{K} \kappa_k \bigl(v_n^{t+1} - v_k^{t+1}\bigr)^2
    \tag{29}\label{eq:33}
\end{gather}
where $\kappa_k>0$ is the coordination coefficient associated with the $k$-th vehicle.

The optimal next state minimizing future coordination cost can be determined by differentiating~\eqref{eq:33} with respect to $v_n^{t+1}$:
\begin{align}
    &\frac{\partial}{\partial v_n^{t+1}}
    \sum_{k=1}^{K} \kappa_k \bigl(v_n^{t+1}-v_k^{t+1}\bigr)^2 \nonumber=0
    \tag{30}\label{eq:34}
\end{align}
which yields the unique minimizer
\begin{gather}
    v_{n,\star}^{t+1}
    = \frac{\sum_{k=1}^{K}\kappa_k v_k^{t+1}}
           {\sum_{k=1}^{K}\kappa_k}.
    \tag{31}\label{eq:35}
\end{gather}

We write
$\kappa_k = \mu + \epsilon_k$, where
$\mu := K^{-1}\sum_{k=1}^{K}\kappa_k > 0$ denotes the nominal coordination strength and
$\epsilon_k := \kappa_k - \mu$ captures the residual heterogeneity. Note that, by definition,
$\sum_{k=1}^{K}\epsilon_k = 0$.
Substituting into~\eqref{eq:35} and
$\bar{v}_{l_n^{t+1}} := \frac{1}{K}\sum_{k=1}^{K} v_k^{t+1}$, we obtain the following exact identity:
\begin{align}
    v_{n,\star}^{t+1} - \bar{v}_{l_n^{t+1}}
    &= \frac{1}{K\mu}\sum_{k=1}^{K}\epsilon_k\bigl(v_k^{t+1}-\bar{v}_{l_n^{t+1}}\bigr).
    \tag{32}\label{eq:36}
\end{align}

~\eqref{eq:36} implies that the deviation of $v_{n,\star}^{t+1}$ from the lane mean speed
$\bar{v}_{l_n^{t+1}}$ is driven by the heterogeneity $\{\epsilon_k\}$.
By the Cauchy--Schwarz inequality,
\begin{align}
    \bigl|v_{n,\star}^{t+1}-\bar{v}_{l_n^{t+1}}\bigr|
    &\le \frac{1}{K\mu}
    \sqrt{\sum_{k=1}^{K}\epsilon_k^2}\,
    \sqrt{\sum_{k=1}^{K}\bigl(v_k^{t+1}-\bar{v}_{l_n^{t+1}}\bigr)^2} \nonumber\\
    &= \frac{\sigma_{\kappa}}{\mu}\,\sigma_v
    \tag{33}\label{eq:37}
\end{align}
where $\sigma_{\kappa}^2 := \frac{1}{K}\sum_{k=1}^{K}(\kappa_k-\mu)^2$ and
$\sigma_v^2 := \frac{1}{K}\sum_{k=1}^{K}\bigl(v_k^{t+1}-\bar{v}_{l_n^{t+1}}\bigr)^2$.
Hence, $v_{n,\star}^{t+1}$ is close to $\bar{v}_{l_n^{t+1}}$ whenever $(\sigma_{\kappa}/\mu)\sigma_v$ is small.

Considering the inherent diversity of traffic scenarios, the same vehicle influenced by a neighboring vehicle may experience minimal coordination effects in some scenarios (e.g., a simple lane change), but significant effects in others (e.g., consecutive accelerations to match the speed). To approximate the future coordination cost under this variability, we assume that in expectation over traffic scenarios, the coordination coefficient is approximately uniform across vehicles. Moreover, since the speed is discretized into only five levels, the speed variance $\sigma_v$ remains bounded. Consequently, $\mathbb{E}\!\left[(\sigma_{\kappa}/\mu)\sigma_v\right]$ is small (equivalently, $\mathbb{E}\!\left[\bigl|v_{n,\star}^{t+1}-\bar{v}_{l_n^{t+1}}\bigr|\right]$ is small by~\eqref{eq:37}).

Under this assumption, we have the approximation $\mathbb{E}\!\left[v_{n,\star}^{t+1}\right]\approx \bar{v}_{l_n^{t+1}}$. Since~\eqref{eq:33} is strictly convex in $v_n^{t+1}$ and attains its minimum at $v_{n,\star}^{t+1}$, it increases with the deviation $\bigl|v_n^{t+1}-v_{n,\star}^{t+1}\bigr|$. Based on this property, we approximate:
\begin{gather}
    \mathbb{E}[\mathrm{FCC}_n(t\mid s)]
    \approx h\!\left(\bigl|v_n^{t+1}-\bar{v}_{l_n^{t+1}}\bigr|\right)
    \tag{34}\label{eq:38}
\end{gather}
where $h: \mathbb{R}_{\geq 0} \to \mathbb{R}_{\geq 0}$ is a monotonically increasing function.

\subsubsection{Approximate Equivalence to the Strategy Function}
Substituting~\eqref{eq:38} into~\eqref{eq:32} yields
\begin{gather}
    \mathbb{E}[\text{Contribution}_n(t \mid s)]
    \approx g^t_n(s) + h\bigl(\bigl|v_n^{t+1} - \bar{v}_{l^{t+1}_n}\bigr|\bigr).
    \tag{35}\label{eq:39}
\end{gather}

The strategy function of OV $n$ is designed as $F_n(s) = w_1 f_1(s) + w_2 f_2(s) + w_3 f_3(s)$, where $f_1(s)$ measures the instantaneous behavior changes (lane changes and speed changes), $f_2(s)$ is the absolute speed deviation from the lane mean speed, and $f_3(s)$ is a penalty term enforcing safety and efficiency constraints.

When the candidate next state $s$ satisfies all safety and efficiency constraints, 
we have $f_3(s) = 0$. Let $S_{\text{safe}} := \{\, s \mid f_3(s) = 0\}$
denote the set of candidate next states that satisfy all safety and efficiency constraints.
In this case, the strategy function reduces to
\begin{align}
    F_n(s)
    &= w_1 g^t_n(s) + w_2 \bigl|v_n^{t+1} - \bar{v}_{l^{t+1}_n}\bigr|
    \tag{36}\label{eq:41}
\end{align}
Comparing (\ref{eq:39}) and (\ref{eq:41}), we observe that both expressions share the same instantaneous behavior cost term $g_n^t(s)$. The remaining terms, $h\bigl(\bigl|v_n^{t+1} - \bar{v}_{l^{t+1}_n}\bigr|\bigr)$ in (\ref{eq:39}) and $w_2 \bigl|v_n^{t+1} - \bar{v}_{l^{t+1}_n}\bigr|$ in (\ref{eq:41}), are both monotonically increasing functions of the same argument $\bigl|v_n^{t+1} - \bar{v}_{l^{t+1}_n}\bigr|$.
Therefore, by selecting appropriate weights $w_1$ and $w_2$, $F_n(s)$ and $\mathbb{E}\big[\text{Contribution}_n(t \mid s)\big]$ produce approximately the same ranking over all candidate states $s \in S_{\text{safe}}$, which implies
\[
\arg\min_{s \in S_{\text{safe}}} F_n(s) \approx \arg\min_{s \in S_{\text{safe}}} \mathbb{E}\big[\text{Contribution}_n(t \mid s)\big].
  \tag{37}\label{eq:42}
\]

\section{SIMULATION RESULTS}\label{sec:experiments}

We first describe the experimental setup, including the real-world traffic dataset, discretization method, baseline algorithms, and parameter configurations. We then compare SDVC with existing methods in terms of solution quality and computational efficiency. Finally, we conduct scalability evaluation, including complete route planning, varying traffic density and speed heterogeneity, and varying lane configurations. All experiments are implemented in Python and executed on a computer with an Intel Core i7-10750H processor and 16.0 GB RAM. Gurobi (version 12.0.0) is used as the commercial solver for baseline methods.

\subsection{Experimental Setup}
The data used come from the HighD dataset of German highways~\cite{38}, which contains vehicle trajectory data on six-lane bidirectional highway segments. The road segment is discretized into a grid system (Fig.~\ref{1}), and speeds are discretized according to Table~\ref{tab:speed_lookup}. For experimental samples, we select frames from the 25th data track, where frames 28606, 28716, and 29246 follow the settings in~\cite{30}, and the remaining frames (28024 and 28378) are manually configured with planning time steps set to ensure all EMVs can pass through the road segment. The weight parameters in~\eqref{eq:2} are set to $c_1 = c_2 = c_3 = 1$, consistent with~\cite{30}. Three metrics are used for evaluation: the value of $f'$ defined in~\eqref{eq:2}, total planning time, and collision rate (calculated as the number of vehicles involved in collisions divided by the total number of vehicles during the planning horizon).

The state-of-the-art methods include LPH, CRHA~\cite{30}, and GSAC~\cite{Yang}. To ensure fair comparison, we require real-time vehicle control capability and maintain the same decision frequency as the SDVC algorithm. We select CRHA and GSAC as baselines because they satisfy real-time requirements, while LPH is excluded due to its inability to meet real-time constraints. We further propose two improved variants: CRHA\_1, which applies LPH's spatial optimization strategy only at the beginning, and CRHA\_2, which applies it before each horizon window. Since GSAC is only effective within 100~$m$ ahead of the EMV, we employ the Intelligent Driver Model (IDM)~\cite{IDM} for vehicles beyond this range, consistent with~\cite{Yang}, and refer to this combined approach as GSAC\_IDM.
Tables~\ref{tab:parameters} and~\ref{tab:gsac_parameters} summarize the main parameters for all comparison algorithms. For CRHA and its variants, we introduce the rolling step $\Delta < T_H$ to control planning frequency: when computation time exceeds $\Delta$, the algorithm adopts the current best solution. We evaluate all algorithms across different $(T_H, \Delta)$ combinations.

\begin{figure}[t]
    \centering
    \includegraphics[width=\linewidth]{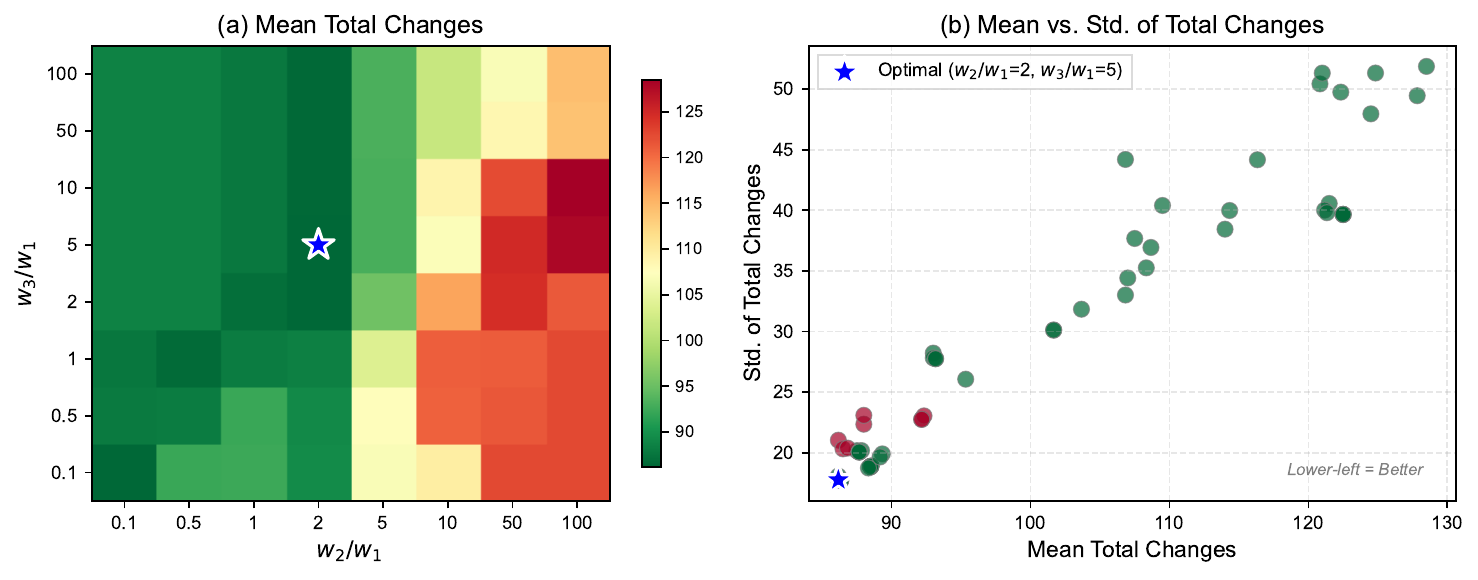}
    \caption{Sensitivity analysis of weight parameters. (a) Mean total lane changes. (b) Mean versus standard deviation. The blue star indicates the optimal configuration.}
    \label{fig:sensitivity}
\end{figure}

In SDVC, the communication range of vehicles is an important parameter. Typically, vehicles equipped with V2V communication devices can communicate within a range of 300-500 $m$~\cite{39}. We select 400 $m$ as our standard communication range. To determine the weight parameters $(w_1, w_2, w_3)$ in the strategy function, we conducted a sensitivity analysis across multiple heterogeneous traffic scenarios. Fixing $w_1 = 1$ as the baseline, we systematically varied the ratios $w_2/w_1$ and $w_3/w_1$ over the range $\{0.1, 0.5, 1, 2, 5, 10, 50, 100\}$, yielding $64$ parameter configurations. 
To balance overall performance and cross-scenario consistency, we defined a composite score $S = \bar{f} + \sigma_f$,
where $\bar{f}$ and $\sigma_f$ denote the mean and standard deviation of total changes, respectively. As shown in Fig.~\ref{fig:sensitivity}, the configuration $(w_1, w_2, w_3) = (1, 2, 5)$ achieved the minimum $S$ among all configurations.

\begin{table}[htbp]
\centering
\caption{MAIN PARAMETERS USED IN COMPARISON ALGORITHMS}
\label{tab:parameters}
\renewcommand{\arraystretch}{1.2}
\resizebox{\columnwidth}{!}{%
\begin{tabular}{@{}cccc@{}}
\toprule
\textbf{Name}& \textbf{Type} & \textbf{Dimension} & \textbf{Domain} \\
\midrule
$T_{H}$ & Integer variable& 1& [3, 12] \\
$\triangle$ & Integer variable& 1& [1, $T_{H}$) \\
$i^t_{n}$ & Integer variable & $(T_{H}, |\mathcal{N}|)$& [1, $I$]\\
$a^{t}_{n}$ & Integer variable & $(T_{H}-1, |\mathcal{N}|)$ & [0, 1] \\
$b^{t}_{n}$ & Integer variable & $(T_{H}-1, |\mathcal{N}|)$ & [0, 1] \\
$l^t_{n}$ & Integer variable & $(T_{H}, |\mathcal{N}|)$ & [1, 3] \\
$l^t_{m}$ & Integer variable & $(T_{H}, |\mathcal{M}|)$ & [1, 3] \\
$v^t_{n}$ & Integer variable & $(T_{H}, |\mathcal{N}|)$ & [1, $V_{max}$]\\
$\alpha^{t}_{n}$ & Binary variable & $(T_{H}-1, |\mathcal{N}|)$ & $\{0, 1\}$ \\
$\beta^{t}_{n}$ & Binary variable & $(T_{H}-1, |\mathcal{N}|)$ & $\{0, 1\}$ \\
$w_{t,n',n}$ & Binary variable & $(T_{H}, |\mathcal{N}|, |\mathcal{N}|)$ & $\{0, 1\}$ \\
$\phi_{t,m,n}$ & Binary variable & $(T_{H}, |\mathcal{M}|, |\mathcal{N}|)$ & $\{0, 1\}$ \\
$M_1$ & Constant & - & $I$\\
$M_2$ & Constant & - & 3$I$\\
\bottomrule
\end{tabular}%
}
\renewcommand{\arraystretch}{1.0}
\end{table}

\begin{table}[htbp]
\centering
\caption{PARAMETER SETTINGS OF GSAC\_IDM}
\label{tab:gsac_parameters}
\renewcommand{\arraystretch}{1.2}
\resizebox{\columnwidth}{!}{%
\begin{tabular}{@{}lc@{}}
\toprule
\textbf{Parameter} & \textbf{Value} \\
\midrule
\multicolumn{2}{l}{\textit{GSAC Training Environment}} \\
Number of lanes & 3 \\
Traffic density & 110 veh/km \\
Average initial OV speed $\bar{V}_{\text{OV}}$ & 2 \\
\midrule
\multicolumn{2}{l}{\textit{GSAC Network Architecture}} \\
FCN Encoder & 2 layers, 32 units each \\
GCN Encoder & 1 layer, $32 \times 32$ \\
Critic/Actor Network & MLP, $256 \times 256 \times 256$ \\
\midrule
\multicolumn{2}{l}{\textit{GSAC Training Hyperparameters}} \\
Batch size & 256 \\
Discount factor $\gamma$ & 0.99 \\
Learning rate (Critic/Actor) & $5 \times 10^{-4}$ \\
Learning rate (Temperature) & $2 \times 10^{-4}$ \\
Training steps & 400{,}000 \\
Action frequency & 1 Hz \\
\midrule
\multicolumn{2}{l}{\textit{GSAC Reward Function Coefficients}} \\
$(w_1, \eta, w_2, w_3, w_4, w_5, w_6)$ & $(0.2, 30, 1.5, 1, -5, -20, -2)$ \\
\midrule
\multicolumn{2}{l}{\textit{IDM Parameters}} \\
Desired speed $v_0$ & Depending on the scenario \\
Maximum acceleration $a$ & 1 (cells/s$^2$) \\
Comfortable deceleration $b$ & 1 (cells/s$^2$) \\
Minimum gap $s_0$ & Defined by (\ref{eq:8})\\
Desired time headway $T$ & 0 (s)\\
Acceleration exponent $\delta$ & 4 \\
Discretization threshold & 0.5 \\
\bottomrule
\end{tabular}%
}
\renewcommand{\arraystretch}{1.0}
\end{table}

\begin{table*}[htbp]
\centering
\renewcommand{\arraystretch}{1.8}
\setlength{\tabcolsep}{3pt}
\caption{COMPARISON OF DIFFERENT ALGORITHMS}
\label{tab:merged_performance_v2}
\resizebox{\textwidth}{!}{
\begin{tabular}{c|ccccc|ccccc|ccccc}
\hline
\multirow{2}{*}{\textbf{Case}} & \multicolumn{5}{c|}{\textbf{The value of $f^{\prime}$}} & \multicolumn{5}{c|}{\textbf{Total planning time ($s$)}} & \multicolumn{5}{c}{\textbf{Collision Rate (\%)}} \\
\cline{2-16}
 & \textbf{SDVC} & \textbf{GSAC\_IDM}& \textbf{CRHA} & \textbf{CRHA\_1}& \textbf{CRHA\_2} & \textbf{SDVC} & \textbf{GSAC\_IDM}& \textbf{CRHA} & \textbf{CRHA\_1}& \textbf{CRHA\_2} & \textbf{SDVC} & \textbf{GSAC\_IDM}& \textbf{CRHA} & \textbf{CRHA\_1}& \textbf{CRHA\_2}\\
\hline
30\_14\_5  & 16 & 27& \underline{\textbf{13}}/22.38/145 & \underline{\textbf{13}}/16.17/59  & \underline{\textbf{13}}/15.60/30 & 0.036 & \underline{\textbf{0.01}}& 1.07/7.92/14.88  & 0.52/4.39/8.90   & 0.49/4.10/10.28  & 0& 0& 0& 0& 0\\
35\_18\_4  & \underline{\textbf{25}} & 51& 27/83.45/205 & \underline{\textbf{25}}/31.00/88  & \underline{\textbf{25}}/29.11/68          & 0.046 & \underline{\textbf{0.01}}& 1.24/12.85/22.12 & 0.92/11.30/20.07 & 0.58/9.79/20.05  & 0& 0& 0& 0& 0\\
43\_15\_5  & 28 & 58& \underline{\textbf{27}}/96.89/253 & 28/52.81/133 & 28/36.59/49          & 0.036 & \underline{\textbf{0.01}}& 2.04/12.71/18.13 & 1.38/11.52/20.05 & 0.82/10.56/18.09 & 0& 0& 0& 0& 0\\
47\_18\_4  & \underline{\textbf{25}} & 61& 28/100.67/292& 32/54.91/175 & 31/51.70/152         & 0.043 & \underline{\textbf{0.01}}& 2.46/13.91/20.92 & 2.15/12.03/16.49 & 1.19/11.71/20.08 & 0& 6& 0& 0& 0\\
54\_24\_3  & \underline{\textbf{14}} & 24& \underline{\textbf{14}}/28.67/233 & \underline{\textbf{14}}/19.82/55  & \underline{\textbf{14}}/18.82/27          & 0.061 & \underline{\textbf{0.02}}& 2.65/15.85/24.81 & 1.51/9.61/17.55  & 1.48/8.68/15.38  & 0& 0& 0& 0& 0\\
131\_42\_5 & \underline{\textbf{96}} & 144& $\sim$       & $\sim$       & $\sim$               & 0.11  & \underline{\textbf{0.03}}& $\sim$           & $\sim$           & $\sim$           & 0& 3& $\sim$ & $\sim$ & $\sim$ \\
139\_53\_4 & \underline{\textbf{77}} & 188& $\sim$       & $\sim$       & $\sim$               & 0.14  & \underline{\textbf{0.04}}& $\sim$           & $\sim$           & $\sim$           & 0& 7& $\sim$ & $\sim$ & $\sim$ \\
166\_72\_3 & \underline{\textbf{52}} & 78& $\sim$       & $\sim$       & $\sim$               & 0.19  & \underline{\textbf{0.05}}& $\sim$           & $\sim$           & $\sim$           & 0& 1& $\sim$ & $\sim$ & $\sim$ \\
\hline
\multicolumn{16}{l}{Note: "$\sim$" indicates no effective results obtained.} \\
\end{tabular}
}
\end{table*}

\subsection{Experimental Result Analysis}

Table~\ref{tab:merged_performance_v2} summarizes the performance of SDVC against baseline methods across five short-horizon single-segment scenarios. The case identifier ``N\_T\_V'' denotes the number of OVs, planning horizon, and maximum speed level, respectively. For CRHA-based methods, results are reported as best/mean/worst values over all $(T_H, \Delta)$ configurations.

\begin{figure*}[!t]
    \vspace{-2mm} 
    \centering
    \includegraphics[width=\textwidth,
                     trim=0 150 0 140,clip]{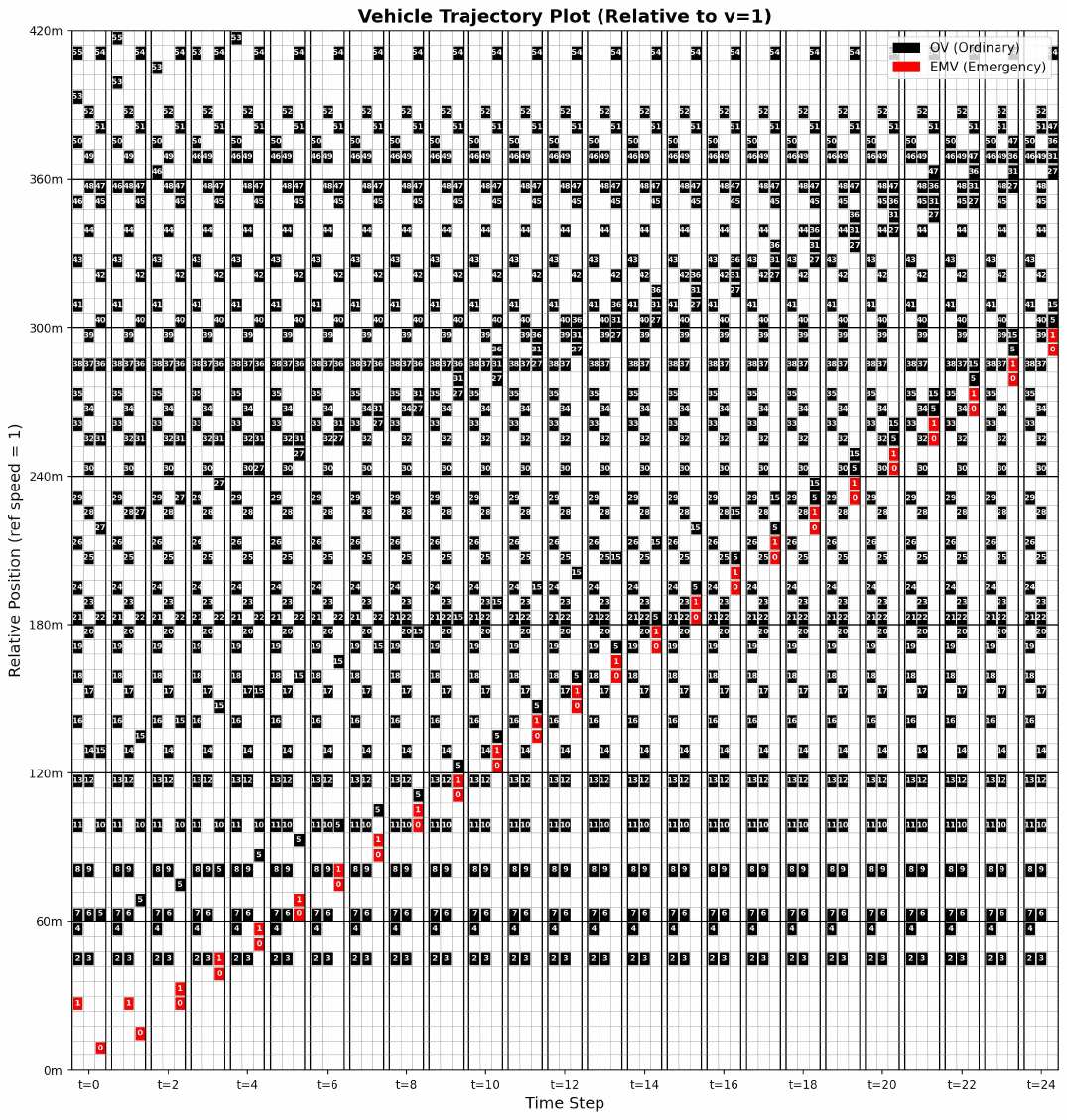}
    \vspace{-3mm} 
    \caption{Vehicle trajectories under the proposed SDVC in case 54\_24\_3.}
    \vspace{-2mm} 
    \label{fig:trajectory}
\end{figure*}

By analyzing the data in Table \ref{tab:merged_performance_v2}, we first compare SDVC with CRHA-based methods. When they do not have prior knowledge, which means that the optimal $T_{H}$ and $\triangle$ are unknown, they can achieve performance comparable to SDVC in small scale scenarios (such as 30\_14\_5). However, as the complexity of the problem increases, SDVC substantially outperforms all CRHA-based methods. When they have prior knowledge, which means using optimized fixed parameter settings determined through extensive experiments as in \cite{30}, SDVC maintains its advantage: it achieves the same result of 28 in case 43\_15\_5, outperforms their result of 30 with 25 in case 47\_18\_4, and outperforms their result of 15 with 14 in case 54\_24\_3. It is worth noting that when compared with the best values obtained by various parameter configurations, the performance of SDVC remains competitive. In cases 54\_24\_3 and 35\_18\_4, it achieves the same best values of 14 and 25, respectively; in case 47\_18\_4, it even outperforms the best value of 28. In the remaining cases, the differences between SDVC and the best values of its peers are minimal. These results demonstrate that SDVC achieves strong performance even in short-horizon single-segment scenarios. Compared to GSAC\_IDM, SDVC reduces $f'$ by 37.0\%--59.0\% across all five cases. 
This improvement can be attributed to the inherent generalization limitations of reinforcement learning-based methods. 
The learned policy still struggles to generalize across diverse real-world scenarios with locally varying traffic densities and heterogeneous vehicle speeds as the EMV advances,
leading to degraded performance and non-zero collision rates. In contrast, SDVC's training-free design naturally adapts to varying traffic conditions while maintaining zero collision rate across all cases. Regarding computational efficiency, SDVC achieves total planning times of 36--61~ms, with per-step latencies well below the human visual reaction time of 200~ms~\cite{40}, fully meeting real-time requirements. Fig.~\ref{fig:trajectory} shows vehicle    trajectories under the proposed SDVC in case 54\_24\_3.

\subsection{Scalability Evaluation}
\label{subsec:scalability}

\subsubsection{Complete Route Planning}

We first conduct complete route planning experiments in long-horizon multi-segment scenarios. Unlike previous experiments that focus on a 420~m road segment, these scenarios involve complete transit routes composed of multiple consecutive segments, such as real-world routes from starting points to hospitals, which better reflect the complexity of actual application environments. To construct these test scenarios, we select specific frame sequences from the 25th data track and connect them sequentially. Frames 28606, 28607, and 28608 form the first test case ``131\_42\_5''. Frames 28716, 28717, and 28718 form the second test case ``139\_53\_4''. Frames 29246, 29247, and 29248 form the third test case ``166\_72\_3''. For CRHA-based methods, to ensure that all vehicles in the scenarios can plan complete routes within the specified planning time steps, $I$ is set to 400.

The last three rows of Table~\ref{tab:merged_performance_v2} show the performance results of SDVC in complete route planning experiments. In these large-scale scenarios, CRHA-based methods fail to yield any effective results due to the exponential growth of decision variables and constraints. In contrast, SDVC successfully plans complete routes in all test scenarios, achieving $f^{\prime}$ values of 96, 77, and 52 in cases 131\_42\_5, 139\_53\_4, and 166\_72\_3, respectively. This feasibility stems from the distributed vehicle control mechanism of SDVC, where each vehicle only utilizes information from vehicles within its communication range, independent of the total number of vehicles in the scenario. Compared to GSAC\_IDM, SDVC reduces $f^{\prime}$ by 33.3\%, 59.0\%, and 33.3\% in these three cases, respectively. Furthermore, SDVC also maintains per-step latencies well below the human visual reaction time of 200~ms, fully meeting real-time requirements.

\subsubsection{Traffic Density and Speed Heterogeneity}
\label{subsubsec:density_speed}
We conduct experiments across different traffic densities and speed heterogeneity levels. The traffic density is defined as the number of OVs per unit highway segment length ($\mathrm{veh/km}$), and speed heterogeneity is characterized by $\Delta v = V_{\text{max}} - \bar{V}_{\text{OV}}$, representing the speed difference between the EMV target speed and the average initial speed of OVs. We select 7 density levels ranging from $64$ to $162~\mathrm{veh/km}$, covering the entire range available in the HighD dataset to ensure comprehensive evaluation under real-world traffic conditions. For each density, we randomly select 3 frames from the 25th data track to construct a long route, with 5 such routes for each density--speed combination, yielding 90 long routes with different configurations in total. The reported results are averaged over these runs. The planning horizon is uniformly set to $72~\mathrm{s}$ to ensure vehicles can traverse the long route and verify algorithm stability.

\begin{figure}[t]
    \centering
    \includegraphics[width=0.85\linewidth]{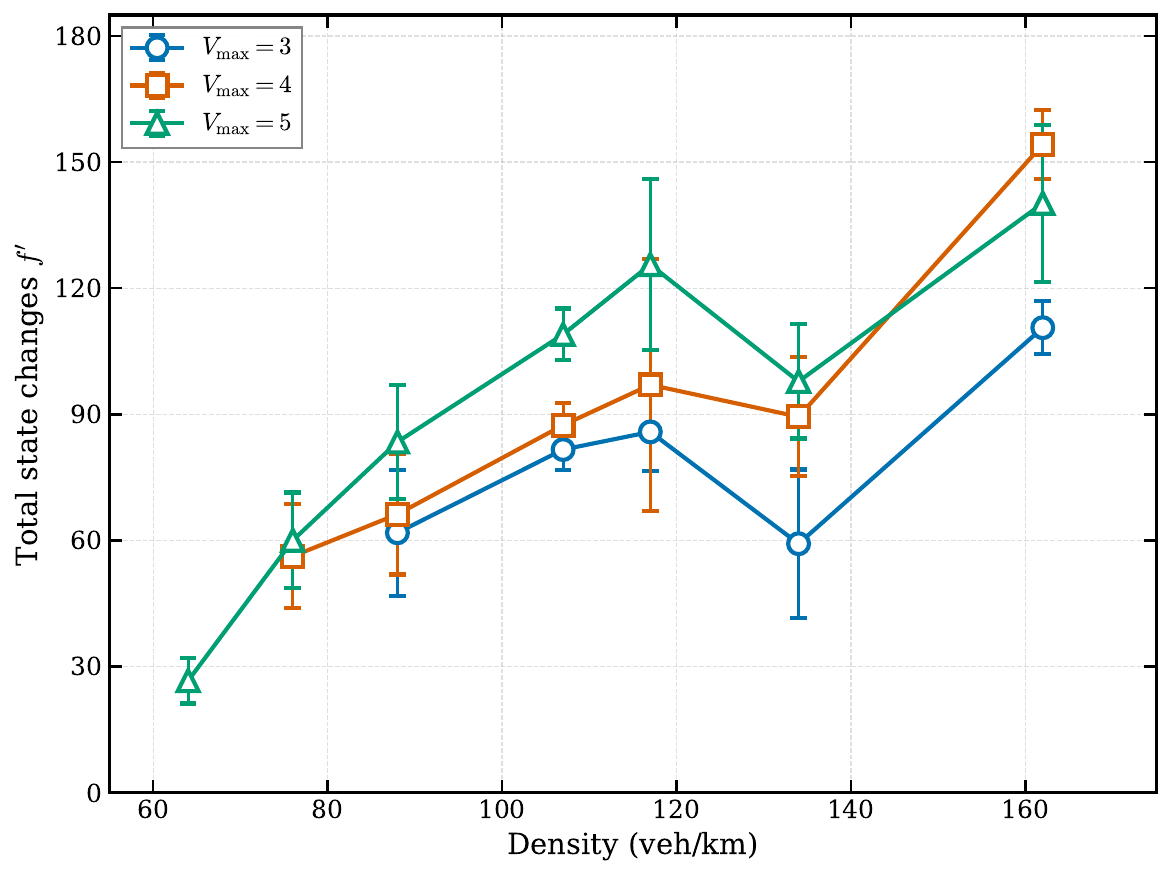}
    \caption{Variation trends of $f'$ under different traffic densities and speed heterogeneity levels.}
    \label{fig:varying}
\end{figure}

The results are recorded in Table~\ref{tab:density_speed}, and Fig.~\ref{fig:varying} illustrates the variation trends. Across the $64$--$134~\mathrm{veh/km}$ density range, the value of $f'$ exhibits a pattern consistent with the optimal solutions obtained by Lin et al.~\cite{30} using mathematical solvers: in low-density scenarios, $f'$ grows slowly as density increases; in medium-density scenarios, $f'$ becomes more sensitive to density growth; in high-density scenarios, $f'$ decreases as density increases due to reduced speed dispersion among vehicles. This consistency validates that SDVC achieves near-optimal performance through distributed vehicle control. For the extremely high density ($162~\mathrm{veh/km}$) that Lin et al.~\cite{30}\ did not consider, we observe that $f'$ increases substantially because OVs in the EMV's target lane have no feasible positions in adjacent lanes for lane-changing, forcing them to accelerate multiple times to match the EMV's speed to avoid collisions.
Moreover, higher speed heterogeneity implies that more vehicles need to change their states to clear lanes for EMVs. The results confirm this: for instance, at $117~\mathrm{veh/km}$, $f'$ increases from $85$ ($\Delta v = 1$) to $125$ ($\Delta v = 3$), a $47.1\%$ increase.
Compared to GSAC\_IDM, SDVC achieves lower $f'$ values and maintains zero collision rate across all configurations, while GSAC\_IDM incurs collision rates of up to 11\% due to limited generalization capability under varying traffic conditions.

\begin{table}[t]
\centering
\caption{Performance Comparison Under Varying Traffic Density and Speed Heterogeneity}
\label{tab:density_speed}
\scriptsize
\renewcommand{\arraystretch}{1.5}
\resizebox{\columnwidth}{!}{%
\begin{tabular}{c|c|c|cc|cc|cc}
\toprule
\multirow{2}{*}{\shortstack[c]{Density\\(veh/km)}} &
\multirow{2}{*}{$|N|$} &
\multirow{2}{*}{$\Delta v$} &
\multicolumn{2}{c|}{$f' \downarrow$} &
\multicolumn{2}{c|}{Collision Rate (\%) $\downarrow$} &
\multicolumn{2}{c}{Decision Time (ms) $\downarrow$} \\
\cline{4-9}
 &  &  & SDVC & GSAC\_IDM& SDVC & GSAC\_IDM& SDVC & GSAC\_IDM\\
\midrule
\multirow{1}{*}{64} & \multirow{1}{*}{81}  & 1 & \underline{\textbf{27}}& 34& \underline{\textbf{0}}& \underline{\textbf{0}}& 2.5& \underline{\textbf{0.75}}\\
\midrule
\multirow{2}{*}{76} & \multirow{2}{*}{96}  & 1 & \underline{\textbf{56}}& 71& \underline{\textbf{0}}  & \underline{\textbf{0}}& 2.5& \underline{\textbf{0.75}}\\
                    &                      & 2 & \underline{\textbf{60}}& 90& \underline{\textbf{0}}  & 1& 2.4& \underline{\textbf{0.75}}\\
\midrule
\multirow{3}{*}{88} & \multirow{3}{*}{111} & 1 & \underline{\textbf{61}}& 126& \underline{\textbf{0}}  & 2& 2.4& \underline{\textbf{0.75}}\\
                    &                      & 2 & \underline{\textbf{66}}& 178& \underline{\textbf{0}}  & 4& 2.5& \underline{\textbf{0.75}}\\
                    &                      & 3 & \underline{\textbf{83}}& 188& \underline{\textbf{0}}  & 4& 2.4& \underline{\textbf{0.75}}\\
\midrule
\multirow{3}{*}{107} & \multirow{3}{*}{129} & 1 & \underline{\textbf{82}}& 126& \underline{\textbf{0}} & 2& 2.5& \underline{\textbf{0.75}}\\
                     &                      & 2 & \underline{\textbf{87}}& 155& \underline{\textbf{0}} & 3& 2.5& \underline{\textbf{0.75}}\\
                     &                      & 3 & \underline{\textbf{109}}& 171& \underline{\textbf{0}} & 3& 2.5& \underline{\textbf{0.75}}\\
\midrule
\multirow{3}{*}{117} & \multirow{3}{*}{141} & 1 & \underline{\textbf{85}}& 158& \underline{\textbf{0}} & 3& 2.6& \underline{\textbf{0.75}}\\
                     &                      & 2 & \underline{\textbf{97}}& 212& \underline{\textbf{0}} & 4& 2.7& \underline{\textbf{0.75}}\\
                     &                      & 3 & \underline{\textbf{125}}& 245& \underline{\textbf{0}} & 6& 2.7& \underline{\textbf{0.75}}\\
\midrule
\multirow{3}{*}{134} & \multirow{3}{*}{168} & 2 & \underline{\textbf{59}}& 101& \underline{\textbf{0}} & 2& 2.5& \underline{\textbf{0.75}}\\
                     &                      & 3 & \underline{\textbf{89}}& 121& \underline{\textbf{0}} & 6& 2.6& \underline{\textbf{0.75}}\\
                     &                      & 4 & \underline{\textbf{98}}& 126& \underline{\textbf{0}} & 5& 2.7& \underline{\textbf{0.75}}\\
\midrule
\multirow{3}{*}{162} & \multirow{3}{*}{204}& 2& \underline{\textbf{111}}& 202& \underline{\textbf{0}}& 8& 2.7& \underline{\textbf{0.75}}\\
                     &                      & 3& \underline{\textbf{154}}& 249& \underline{\textbf{0}}& 8& 2.8& \underline{\textbf{0.75}}\\
                     &                      & 4& \underline{\textbf{149}}& 264& \underline{\textbf{0}}& 11& 3.1& \underline{\textbf{0.75}}\\
\bottomrule
\end{tabular}%
}
\\[2pt]
\raggedright\footnotesize
\end{table}

\subsubsection{Roadway Lane Configurations}
\label{subsubsec:lane_config}

We conduct experiments under varying lane configurations. Existing methods are exclusively validated on 3-lane scenarios, despite the prevalence of multi-lane roads in real-world infrastructures~\cite{five_lane}. This experiment addresses this gap by testing on 3-lane, 4-lane, and 5-lane roadways.

We evaluate both SDVC and GSAC\_IDM under 3-lane, 4-lane, and 5-lane configurations. The 4-lane and 5-lane scenarios are constructed by expanding the 3-lane setup while preserving the same per-lane density, with the number of OVs increased proportionally. The planning horizon is set sufficiently long to ensure all EMVs can traverse the road segment.

The results in Table~\ref{tab:lane_config} demonstrate that SDVC maintains stable performance across all lane configurations, achieving $f'$ values of 125, 156, and 150 for 3-lane, 4-lane, and 5-lane scenarios, respectively, with zero collision rate in all cases. In contrast, GSAC\_IDM exhibits significantly higher $f'$ values (245--304) and non-zero collision rates (4\%--7\%), as the trained model cannot generalize to other lane configurations. These results highlight the superior generalizability of SDVC in handling diverse roadway infrastructures.

\begin{table}[t]
\centering
\caption{Performance Comparison Under Varying Lane Configurations}
\label{tab:lane_config}
\scriptsize
\renewcommand{\arraystretch}{1.5}
\resizebox{\columnwidth}{!}{%
\begin{tabular}{c|c|cc|cc|cc}
\toprule
\multirow{2}{*}{\shortstack[c]{Lane\\Count}} &
\multirow{2}{*}{$|N|$} &
\multicolumn{2}{c|}{$f' \downarrow$} &
\multicolumn{2}{c|}{Collision Rate (\%) $\downarrow$} &
\multicolumn{2}{c}{Decision Time (ms) $\downarrow$} \\
\cline{3-8}
 &  & SDVC & GSAC\_IDM& SDVC & GSAC\_IDM& SDVC & GSAC\_IDM\\
\midrule
3 & 141&  \underline{\textbf{125}}&  245& \underline{\textbf{0}} &  6&  2.7&  \underline{\textbf{0.75}}\\
4 & 192&  \underline{\textbf{156}}& 304& \underline{\textbf{0}} & 7&  2.7& \underline{\textbf{0.75}}\\
5 & 248&  \underline{\textbf{150}}& 288& \underline{\textbf{0}} & 4&  2.7& \underline{\textbf{0.75}}\\
\bottomrule
\end{tabular}%
}
\\[2pt]
\raggedright\footnotesize
Note: Per-lane density is fixed.
\end{table}

\section{CONCLUSION}\label{sec:conclusion}

This paper has presented a scalable distributed vehicle control (SDVC) method for rapid transit of emergency vehicles. The method consists of two key components: a local strategy function that enables each vehicle to make optimal decisions using only local information, while we theoretically show that minimizing it is approximately equivalent to minimizing the expected global optimization objective, and a distributed conflict resolution mechanism that ensures vehicle safety through coalition formation.
Simulation experiments based on real-world traffic datasets validate the effectiveness of SDVC. Compared with the state-of-the-art methods CRHA-based methods and GSAC\_IDM, SDVC achieves superior solution quality while maintaining zero collision rate across all tested scenarios, including short-horizon single-segment scenarios, long-horizon multi-segment scenarios, varying traffic density and speed heterogeneity scenarios, and varying lane configuration scenarios. The results demonstrate that SDVC provides strong scalability and real-time performance that existing methods cannot achieve.
Our future work intends to extend this method to more complex traffic scenarios, such as intersections, and explore collaborative mechanisms with intelligent traffic signal control systems.

\bibliography{Reference}
\bibliographystyle{IEEEtran}

\end{document}